%% file: main.tex
\documentclass{article} 
\usepackage{iclr2027_conference,times}

\input{math_commands.tex}

\usepackage{hyperref}
\usepackage{url}

\usepackage{amssymb}
\usepackage{multirow}
\usepackage{makecell}
\usepackage{graphicx}
\usepackage[table]{xcolor}
\usepackage{svg}
\usepackage{enumitem}
\usepackage{tablefootnote}
\usepackage{amsmath}
\usepackage{booktabs}
\usepackage[dvipsnames]{xcolor}         
\usepackage{soul}
\usepackage{colortbl}
\newcommand{\meanstd}[2]{%
    $#1$\,\scalebox{0.8}{$\pm #2$}%
}

\title{MorphCL:\\Morphological Contrastive Learning for Inertial-based Human Activity Recognition}

\author{%
  \begin{minipage}[t]{\dimexpr\textwidth-2\tabcolsep\relax}
    \centering
    Marius Bock\textsuperscript{1,2,4}\thanks{Corresponding author: \texttt{[lastname]@iai.uni-bonn.de}}\qquad
    Yuwei Zhang\textsuperscript{2}\qquad
    Juergen Gall\textsuperscript{1,4}\\
    Michael Moeller\textsuperscript{3}\qquad
    Kristof Van Laerhoven\textsuperscript{3}\qquad
    Cecilia Mascolo\textsuperscript{2}\\[0.6em]
    \normalfont
    \textsuperscript{1}University of Bonn\qquad
    \textsuperscript{2}University of Cambridge\qquad
    \textsuperscript{3}University of Siegen\\
    \textsuperscript{4}Lamarr Institute for Machine Learning and Artificial Intelligence
  \end{minipage}%
}

\iclrfinalcopy 
\begin{document}

\maketitle
\lhead{Preprint}
\begin{abstract}
Despite the ubiquity of sensors in wearable and mobile devices and the abundance of human movement data they generate, translating unlabeled recordings into foundational motion models remains an open challenge. Self-supervised learning (SSL) has alleviated the need for costly annotations, yet existing approaches leave the global structure of large-scale motion data largely untapped, relying on randomly sampled batches and local comparisons that become particularly problematic for in-the-wild inertial data dominated by stationary, low-variance behaviors. Here we introduce \emph{Morphological Contrastive Learning (MorphCL)}, a self-supervised pretraining framework that uses structure-aware grouping to inject explicit modeling of \emph{global structure} into inertial-based SSL approaches. Building on two well-established pillars of motion analysis, the discovery of motion primitives, or \emph{motifs}, and domain-specific feature descriptors, we show that MorphCL substantially improves linear probing and finetuning results of learned encoders by up to 15 percentage points in F1-score. In a comparison with existing foundation models, we demonstrate that MorphCL-pretrained encoders match or surpass them models in linear probing performance while trained on $4600\times$ less data. Qualitative analysis of the resulting embedding spaces further reveals morphologically meaningful cluster structure, with improved separation of kinematically similar activity classes.
\end{abstract}

\begin{figure}[htbp]
    \centering
    \includegraphics[width=.9\linewidth]{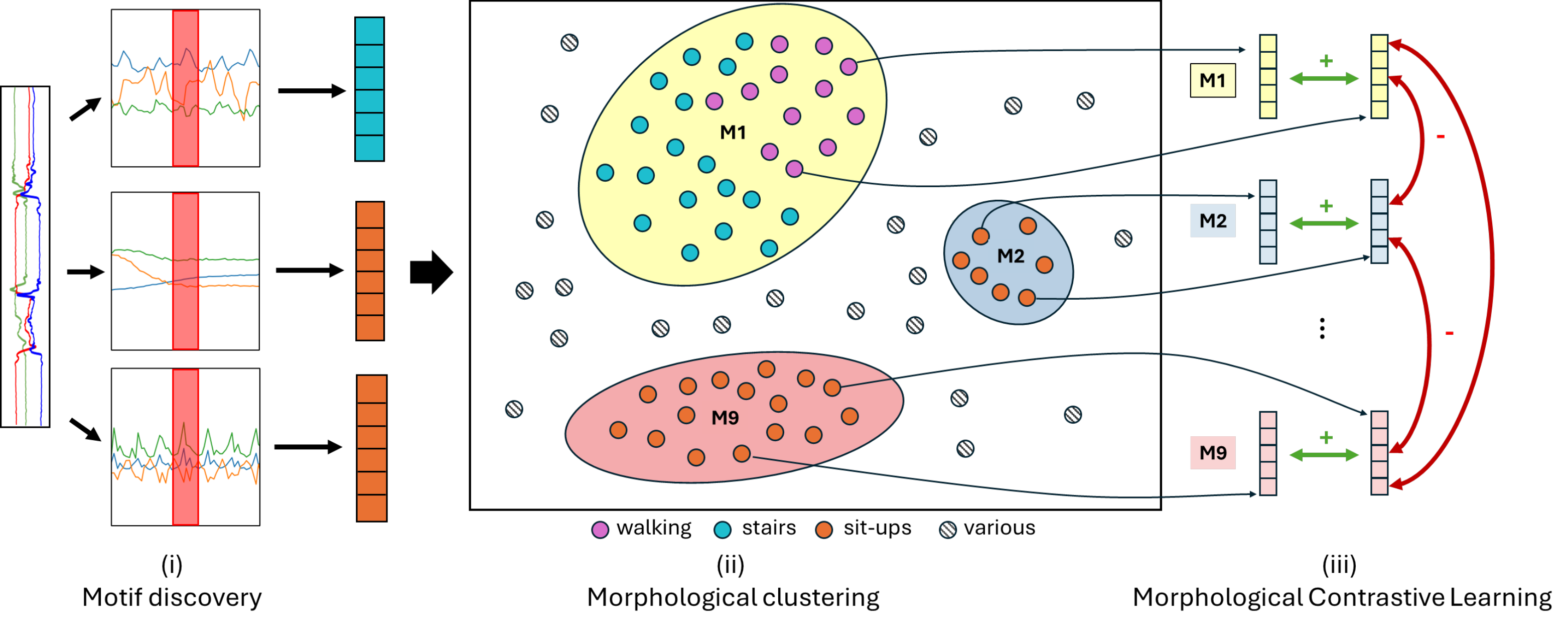}
    \caption{\textbf{Morphological Contrastive Learning (MorphCL)} explicitly models global motion structure by (i) discovering recurring motifs (red) from raw accelerometer data, (ii) clustering them into morphological groups across the pretraining corpus, (iii) defining a group-based contrastive objective that drives agreement beyond instance-level comparisons. \vspace{-5pt} } 
    \label{fig:banner}
\end{figure}

\section{Introduction}
Activities and movements performed during daily life encode rich information about an individual's behavioral patterns, physiological state, and habitual routines~\citep{tryonActivityMeasurement2006}. The automatic identification of such behaviors, commonly referred to as Human Activity Recognition (HAR), has therefore attracted growing interest across a range of health-related research fields~\citep{pattersonFineGrainedActivityRecognition2005, wardActivityRecognitionAssembly2006}. Inertial Measurement Units (IMUs), now ubiquitous in consumer wearables and mobile devices, have emerged as a particularly practical sensing modality for this purpose, yielding discriminative representations well-suited to activity recognition systems~\citep{bullingTutorialHumanActivity2014}.
However, while other domains such as computer vision have seen the rise of large-scale foundation models, which can generalize easily to unseen tasks being trained on vast amounts of data, inertial-based HAR research remains largely confined to small-scale benchmark studies. This gap is driven in part by the difficulty of annotating raw IMU signals retrospectively without additional context, making the construction of labeled datasets~\citep{reissIntroducingNewBenchmarked2012, roggenCollectingComplexActivity2010} costly and predominantly restricted to controlled laboratory environments. Although such datasets have enabled verification of methodological progress on relatively small neural networks~\citep{ordonezDeepConvolutionalLSTM2016}, they lack the scale and diversity required to train modern architectures and fail to capture the variability of real-world human behavior.

The introduction of self-supervised learning (SSL) to inertial HAR has presented a paradigm shift. By no longer requiring annotations of human activities, it has unlocked the use of the large corpora of unlabeled data, generated by everyday wearables.
While cleverly crafted pretext tasks such as contrasting differently augmented views of input signals have shown to perform well on various downstream tasks~\citep{tangExploringContrastiveLearning2021}, such methods typically rely on instance-level discrimination, treating each segment as independent. However, this assumption is often violated in IMU data, where recurring motion primitives are shared across activities and subjects, making instance-level objectives poorly aligned with the underlying structure~\citep{xuREBARRetrievalbasedReconstruction2024}.

Recent work relaxes this constraint by defining positives beyond simple augmentations. Methods like REBAR and RelCon~\citep{xuREBARRetrievalbasedReconstruction2024, xuRelConRelativeContrastive2025} retrieve similar segments across a timeseries and introduce ranking-based objectives. While improving pair selection, they do not use information about dataset composition and still rely on randomly sampled batches. This becomes particularly problematic in large-scale, in-the-wild datasets, which are highly imbalanced and dominated by repetitive, low-variance activities. For example, in the largest open dataset CAPTURE-24~\citep{chanCAPTURE24LargeDataset2024}, approximately 75\% of the data corresponds to stationary behaviors, while more informative activities are rare. As a result, randomly sampled batches are likely to contain many near-identical segments, sustaining the false negative problem~\citep{chuangDebiasedContrastiveLearning2020} at the batch level even with  well-chosen pairs. This can lead to weak training signals and limits contrastive objectives from learning discriminative representations (as shown in Section~\ref{ch:results}).

Based on this observation, we argue that a key missing component in current SSL approaches for HAR is the explicit modeling of \emph{global structure} in motion data. Human movement exhibits recurring patterns, such as periodic walking cycles or characteristic transitions, that appear across time, activities, and individuals~\citep{minnenDiscoveringCharacteristicActions2006}. The automatic discovery of these \emph{motifs} has long served as a fundamental concept in motion analysis, acting as atomic building blocks for recognizing complex activities~\citep{minnenDiscoveringCharacteristicActions2006, hamidStructureStatisticsUnsupervised2007}. We therefore introduce \emph{Morphological Contrastive Learning (MorphCL)}, a self-supervised pretraining framework that can serve as a plug-in solution to inject explicit modeling of \emph{global structure} into existing SSL approaches (see Figure~\ref{fig:banner}). By combining motif discovery with domain-specific feature descriptors, MorphCL constructs a morphological clustering of the pretraining corpus and uses the resulting grouping to define a group-based contrastive objective that encourages agreement between segments sharing similar motion patterns.
Our contributions are three-fold:

\begin{itemize}[leftmargin=*]
    \item We propose a novel morphological clustering system for accelerometer data, which uses a combination of motif-discovery and domain-specific precomputed features to extract morphological structures within an unlabelled pretraining corpus.
    \item Using the extracted clusters, we devise Morphological Contrastive Learning (MorphCL), a group-based contrastive objective that optimizes agreement between windows with similar motion morphology, moving beyond instance-level comparisons.
    \item We demonstrate that including explicit modeling of morphological structure can improve downstream results if combined with existing SSL frameworks (up to 15\% in F1-score) and yields representations competitive with a foundation model that was trained on $4600\times$ more data~\citep{yuanSelfsupervisedLearningHuman2024}, suggesting that domain-informed inductive bias can complement large scale training.
\end{itemize}
    
\section{Related Work}
\textbf{Inertial-based Contrastive Learning.}
Contrastive learning has become a prominent SSL approach for addressing the limited availability of annotated data in Human Activity Recognition~\citep{haresamudramAssessingStateSelfSupervised2022, tangExploringContrastiveLearning2021}. Methods such as SimCLR~\citep{tangExploringContrastiveLearning2021} follow an \textit{instance-discrimination} setup, where positive pairs are created by augmenting the same temporal window (e.g., jittering, scaling, permutation) and all other samples in the batch are treated as negatives.
While effective, this formulation assumes that augmentations sufficiently capture semantic similarity, which is often not the case for complex and noisy motion signals~\citep{xuREBARRetrievalbasedReconstruction2024}.
To overcome the rigidity of augmentation-defined positives, recent work has explored more flexible similarity measures.  
REBAR~\citep{xuREBARRetrievalbasedReconstruction2024} shifts away from purely temporal or augmentation-defined positives by using a learned reconstruction error as a proxy for semantic similarity, and RelCon~\citep{xuRelConRelativeContrastive2025} introduces a relative ranking loss to preserve hierarchical relationships between motions, avoiding the limitations of rigid positive-negative binaries.
However, these approaches remain \emph{instance-centric} and depend on \emph{randomly sampled batches}, without explicitly modeling global structure in motion data. In contrast, our work introduces a structure-aware, group-based objective that captures recurring morphological patterns across large-scale, in-the-wild datasets.

\textbf{Group-based Contrastive Clustering.}
Several studies have shown that incorporating group-level structure can improve contrastive learning. Supervised Contrastive Learning (SupCon)~\citep{khoslaSupervisedContrastiveLearning2020} extends the contrastive framework by defining positives as all samples sharing the same label, leading to more stable and semantically meaningful representations. A parallel line of work jointly optimizes cluster assignments and representations~\citep{ymSelflabellingSimultaneousClustering2020, vangansbekeSCANLearningClassify2020, caronUnsupervisedLearningVisual2020}. These methods were developed primarily for vision, where grouping is typically inferred from raw inputs or learned embeddings. In sensor-based HAR, however, domain features are rich, physically grounded, and known to be discriminative across a wide range of activities~\citep{vanderdoncktNotSleepTraditional2023}. Closest to our approach, PaPaGei~\citep{pillaiPaPaGeiOpenFoundation2025} introduced a domain-specific proxy (sVRI) to define groups of related samples without explicit labels in the context of Photoplethysmogram (PPG) signals. MorphCL extends this intuition to inertial data, where the higher variability and multi-axis dynamics prevent the use of a single scalar proxy, instead combining motif discovery with a broader domain feature set to construct morphological groups. MorphCL decouples clustering from representation learning entirely, providing a plug-in solution that is semantically grounded and complements rather than replaces existing instance-level objectives. To our knowledge, no prior SSL framework for inertial data has explicitly leveraged such structure-aware grouping to inject domain knowledge into the contrastive objective.

\textbf{Foundation Models for Motion Data.}
Recent efforts have begun scaling SSL approaches to large wearable datasets to train foundation models for motion data. \citet{narayanswamy2025scaling} and \citet{zhangSensorLMLearningLanguage2025} train wearable foundation models using minute-level handcrafted features instead of raw signals, which, however, makes them less applicable for temporally precise motion tasks such as HAR. \citet{yuanSelfsupervisedLearningHuman2024} use 700k person-days of data from UK Biobank to train an inertial representation model. SlotFM~\citep{parkSlotFMMotionFoundation2025} introduces a slot-based attention mechanism to decompose motion signals into disentangled latent components. 
Among these models, only \citet{yuanSelfsupervisedLearningHuman2024} open-source model weights so we include a comparison with the model.

\section{Methodology}
\label{sec:method}
We propose Morphological Contrastive Learning (MorphCL), a self-supervised pretraining framework for inertial time series that injects dataset-level global structure into existing SSL approaches. Building on two well-established pillars of motion analysis, MorphCL constructs morphology-aware pseudo-labels from unlabeled motion data and uses these labels to define positives beyond standard augmentation pairs. As shown in Figure~\ref{fig:banner}, MorphCL consists of three stages: (i) motif discovery, (ii) clustering of domain-specific features, and (iii) a group-based contrastive objective. We describe each component below.

\subsection{Motif-based Pre-filtering}
\label{sec:motiffil}
Let
\begin{equation}
\mathbf{T} = \{ \mathbf{x}_t \}_{t=1}^{N}, \quad \mathbf{x}_t \in \mathbb{R}^{D}
\end{equation}
denote a multivariate inertial time series of length $N$ with $D$ sensor axes. Prior work in activity recognition has long viewed complex activities as compositions of shorter repeated patterns, or \emph{motifs}~\citep{minnenDiscoveringCharacteristicActions2006, hamidStructureStatisticsUnsupervised2007}. We thus use motif discovery to identify prominent repeated patterns ("motion primitives") within $\mathbf{T}$ and extract their occurrences. Since activities are typically expressed through coordinated motion across axes~\citep{yehMatrixProfileVI2017}, we perform motif discovery using \emph{multi-dimensional matrix profiling} as proposed by~\citet{yehMatrixProfileVI2017} rather than analyzing each channel independently. Since the full time series can be very large, we perform motif discovery independently on temporally contiguous chunks, identifying up to $m$ distinct motif patterns per chunk and extracting $r$ instances of each which do not exceed a pre-defined motif distance threshold $\sigma$, as measured by the matrix profiling. This way we identify subsequences that are simultaneously similar across a subset of sensor axes. Specifically, we require each motif to be supported by at least two of the three accelerometer axes.

Since different motifs may occur with highly varying frequencies, na\"ively selecting $r$ reoccurences of motifs can result in many near-duplicate patterns originating from a single, frequently occurring activity. To promote diversity among the discovered motifs, we adopt a greedy masking strategy, masking all reoccurrences of a motif that do not exceed $\sigma$ within the current chunk before discovering the next most prominent motif. This prevents repeatedly selecting highly similar instances of the same pattern and encourages the discovery of motifs corresponding to a broader range of activities and motion primitives.

Let
\begin{equation}
\mathcal{I} = \{ i_1, \dots, i_M \} \subset \{ 1, \dots, N \}
\end{equation}
denote the center indices of the $M$ motif occurrences collected across all chunks.

For each $i \in \mathcal{I}$, we extract a \emph{motif window} $\mathbf{w}_i = \{x_{i-W/2}, x_{i+1-W/2} ..., x_{i+W/2}\}^T$ and a \emph{context window} $\mathbf{c}_i = \{x_{i-C/2}, x_{i+1-C/2} ..., x_{i+C/2}\}^T$, given a predetermined window size $W$ and context window size $C$ ($C \ge W$). To avoid redundancy across motifs, we retain only motif windows that do not overlap more than 50\% with any other motif window. The union of all retained motif windows ($\mathcal{W} = \{ \mathbf{w}_i \mid i \in \mathcal{I} \}$),  and context windows ($\mathcal{C} = \{ \mathbf{c}_i \mid i \in \mathcal{I} \}$), 
forms the input to the subsequent morphological clustering stage. 

\subsection{Morphological Clustering}
\label{sec:morphcl}

Having identified repeated patterns in $\mathbf{T}$, we next group these patterns according to their morphology. Unlike physiological signals such as Photoplethysmogram (PPG), which can often be summarized using a small number of domain-specific proxy measures (e.g., sVRI~\citep{pillaiPaPaGeiOpenFoundation2025}), accelerometer data exhibit highly variable, multi-axis dynamics that cannot be captured by a single descriptor~\citep{vanderdoncktNotSleepTraditional2023}. 
We therefore represent each context window $\mathbf{c}_i \in \mathcal{C}$ as a feature vector $\mathbf{z}_i = \Phi(\mathbf{c}_i) \in \mathbb{R}^{F}$ where $\Phi(\cdot)$ concatenates statistical features commonly used in classical machine-learning-based activity recognition. Specifically, we reuse the feature set described by~\citep{vanderdoncktNotSleepTraditional2023} as it has shown to provide strong discriminative capabilities in describing numerous activities in inertial-based data reliably for downstream classification purposes~\citep{vanderdoncktNotSleepTraditional2023, vanderdoncktLeftRightSwappingUpperLower2024, vanderdoncktMagnitudeRotationInvariant2024}.

Since for example flat-line windows may lead to undefined feature values, we further postprocess $\Phi(\mathbf{c}_i)$ by replacing undefined values with 0 or an appropriate upper/lower bound. The details are in the Appendix. Next, we cluster the set of extracted morphological feature vectors $\mathcal{Z} = \{ \mathbf{z}_i \mid i \in \mathcal{I} \}$ using HDBSCAN~\citep{mcinnesHdbscanHierarchicalDensity2017}, a density-based clustering algorithm well suited for clusters of varying size and the presence of outliers. Formally, HDBSCAN induces a mapping 
\begin{equation}
\kappa : \mathbb{R}^{F} \rightarrow \{ 1, \dots, K \} \cup \{ \varnothing \}
\end{equation}

where $K$ denotes the number of morphological clusters discovered in $\mathcal{Z}$, and $\varnothing$ indicates outliers rejected by the algorithm.
Importantly, the number of clusters $K$ is not fixed a priori but is determined automatically by HDBSCAN based on the density structure of the data and the chosen support parameter $\rho$.
The resulting motif-filtered and morphologically labeled dataset (excluding outliers) 
\begin{equation}
\mathcal{D} = \{ (\mathbf{w}_i, y_i) \mid y_i \neq \varnothing , i\in \mathcal{I}\},
\end{equation}
with $y_i =\kappa(\phi(\mathbf{c}_i))$, thus represents a collection of motif windows, each assigned to one of the $K$ morphological groups, defining which windows are treated as positives (same group) and negatives (different groups) in our pretraining corpus. A sensitivity analysis on chosen clustering parameters can be found in the Appendix.

\subsection{Morphological Contrastive Learning (MorphCL)}

With morphological pseudo-labels at hand, we can now define a contrastive objective, Morphological Contrastive Loss (MorphCL), that exploits dataset-level group structure. Given a batch $\mathcal{B} \subset \mathcal{D}$ of motif windows, we encode each motif window in $\mathcal{B}$ using a feature encoder ($f_\theta$) as well as a projection head ($g_\theta$) resulting in latent vectors $\mathbf{u}_i \in \mathbb{R}^E$. Note that, as in~\citet{khoslaSupervisedContrastiveLearning2020}, we extend $\mathcal{B}$ by extending it with an augmented version of each window $\mathbf{w}_i$, randomly drawn from a set of augmentations (noise, scaling, rotation, and negation).   
These augmentations are a subset of those proposed by~\citet{tangExploringContrastiveLearning2021}, selected because they preserve our definition of morphology. 

For a given window  $\mathbf{w}_i$ and its corresponding morphological label $y_i$, we define the set of windows considered to be positive and negative instances of $\mathbf{w}_i$ as 

\begin{equation}
    \mathcal{P}(i)=\{p\in\mathcal{B}\setminus\{i\}\mid y_p=y_i\},
    \qquad 
    \mathcal{N}(i)=\{p\in\mathcal{B}\setminus\{i\}\mid y_p \neq y_i\}.
\end{equation}

That is all windows in $\mathcal{B}$ that share (or do not share) the same morphological label as $\mathbf{w}_i$. 

Based on our definition of $\mathcal{P}$ and $\mathcal{N}$, we define our MorphCL loss as a group-based variant of the Supervised Contrastive Loss (SCL)~\citep{khoslaSupervisedContrastiveLearning2020} 
\begin{equation}
\mathcal{L}_{\mathrm{morph}} = \frac{1}{|\mathcal{B}|}
    \sum_{i \in \mathcal{B}}
- \log
\left( 
\frac{
\sum_{p \in \mathcal{P}(i)} \exp \left( S_{ip} \right)
}{
\sum_{j\in\mathcal{B} \setminus \{i\}} \exp \left( S_{ij} \right)
}
\right)
\end{equation}

where $\exp$ is the exponential and $S_{ij} = \frac{\mathrm{sim}(\mathbf{u}_i, \mathbf{u}_j)}{\tau}$ is the pairwise similarity matrix of embeddings $\mathbf{u}_i$ and $\mathbf{u}_j$ using the cosine similarity function $\mathrm{sim}(\cdot)$ and a temperature value $\tau$. Unlike SCL, which averages log-probabilities computed for each positive pair individually, our formulation thus adopts a group-based perspective. Specifically, SCL treats each positive pair $(i,p)$ independently by normalizing over all other samples in the batch, resulting in a sum over positive pairs outside the logarithm, while MorphCL encourages agreement at the level of morphological groups rather than isolated pairs. An ablation study, comparing MorphCL with SCL as well as a group-based formulation of the NTXentLoss as employed in~\citet{pillaiPaPaGeiOpenFoundation2025} can be found in the Appendix.

\section{Experiments}
\label{sec:experiments}

\subsection{Pretraining}
\textbf{Model.} We assess a wide-range of SSL pretraining pretext tasks~\citep{tangExploringContrastiveLearning2021, saeedMultitaskSelfSupervisedLearning2019, haresamudramMaskedReconstructionBased2020, haresamudramRoleFeaturesHuman2019, xuREBARRetrievalbasedReconstruction2024, xuRelConRelativeContrastive2025} both in combination with and without our proposed MorphCL framework. As encoder networks we use the DeepConvLSTM~\citep{ordonezDeepConvolutionalLSTM2016} as well as a repurposed ViT transformer~\citep{kolesnikovImageWorth16x162021} as recently proposed in~\citet{hoddesScalingLawsWearable2025}. Specifications of both architectures can be found in the Appendix. 

\textbf{Datasets.} As pretraining corpora we use the WISDM~\citep{kwapiszActivityRecognitionUsing2011} and CAPTURE-24~\citep{chanCAPTURE24LargeDataset2024} dataset. The WISDM dataset provides accelerometer and gyroscope data collected by a smartphone and smartwatch with a sampling rate of 20Hz of 51 participants doing various activities within an hour-long recording protocol. The CAPTURE-24 dataset provides 24-hour long, in-the-wild accelerometer recordings of a smartwatch sampling at 100Hz of around 150 participants. To ensure comparability among the two corpora, we only use the smartwatch data of the WISDM dataset and resample both datasets to a common sampling rate of 50Hz. 
We preprocess both the WISDM and CAPTURE-24 dataset by shifting a sliding window of size 128 samples (2.56s) over each participant's data stream without any overlap among the windows. Each pretraining corpus is normalized using the global mean and standard deviation across all windows of the train dataset. Further details on hyperparameters and optimization schemes of each pretraining framework are provided in the Appendix.

\textbf{Training.} We apply MorphCL on a subset of the pretraining corpus, $\mathcal{D}$. The relative size of the motif- and morphological subsets of the WISDM and CAPTURE-24 dataset created by applying the MorphCL framework are summarized in the Appendix. 
Unlike other frameworks which draw windows randomly from the entire pretraining corpus or available data of a participant, our framework creates batches by evenly sampling from the $K$ morphological clusters. As cluster sizes differ, we allow windows of smaller clusters to be sampled multiple times during an epoch such that each window within the largest morphological cluster can be sampled exactly once. Similarly, we allow batches during MorphCL training to be sampled multiple times, such that the tested SSL frameworks, which use the full training data, can sample each window exactly once during each epoch. Ablation studies on these design choices can be found in the Appendix.

\subsection{Downstream Tasks}
We chose a total of five popular HAR benchmark datasets for our downstream evaluation (see Table~\ref{tab:downstreamdatasets}). To make sensor information transferable from our pretraining corpora, we chose only the (right) wrist sensor provided in each of the datasets. We further adjust the sampling rate of each dataset to be 50Hz by resampling accordingly. Unlike existing downstream evaluations~\citep{yuanSelfsupervisedLearningHuman2024}, we include more complex activity recognition tasks such as sports~\citep{bockWEAROutdoorSports2024, hoelzemannHangTimeHARBenchmark2023} and process-related tasks~\citep{schollWearablesWetLab2015}, and do not artificially remove background activity classes (NULL-class) of the downstream datasets, as the accurate differentiation between this class and relevant activities is a necessary trait of activity recognition systems for real-world deployment~\citep{bullingTutorialHumanActivity2014}.

\begin{table}
    \centering
    \scriptsize
    \caption{\textbf{Downstream datasets.} \#Sbjs: participants; \#Cls: activities (+1 indicates dataset provides a NULL-class). Recording duration is averaged across participants. Activities are categorized into periodic (recurring, periodic movements), non-periodic (sporadic, often short movements/gestures) and complex activities (sequential combinations of periodic and non-periodic movements).}
    \label{tab:downstreamdatasets}
    \begin{tabular}{llccccc}
    \toprule
        Dataset                                             & Scenario & Activity Types & \#Sbjs & \#Cls & Avg. Recording (min) \\ \midrule
        \citet{bockWEAROutdoorSports2024}               & outdoor sports workouts & periodic, complex & 22 & 18(+1) & 52.52 ($\pm$ 17.89) \\
        \citet{schollWearablesWetLab2015}             & wetlab DNA extraction & (non-)periodic & 22 & 8(+1) & 47.93 ($\pm$7.48) \\
        \citet{sztylerOnBodyLocalizationWearable2016}  & locomotion & periodic & 13 & 8 & 75.84 ($\pm$4.55) \\
        \citet{hoelzemannHangTimeHARBenchmark2023} & basketball practice and game & (non-)periodic & 24 & 5(+1) & 60.59 ($\pm$10.84) \\
        \citet{reissIntroducingNewBenchmarked2012}    & locomotion and ADL & periodic, complex & 9 & 18(+1) & 50.46 ($\pm$17.79) \\
        \bottomrule
    \end{tabular}
\end{table}

We evaluate both linear probing as well as finetuning capabilities of our pretrained backbones. Details on the chosen hyperparameters and optimization parameters can be found in the Appendix. All results mentioned in this paper are reported as fold-averaged Leave-One-Subject-Out (LOSO) cross-validation results, where each participant in the downstream dataset becomes the test set exactly once, and seed-averaged across three runs.

\section{Results}
\label{ch:results}

\subsection{Complementarity to Existing SSL Methods}
\label{subsec:complementarity}

To investigate whether MorphCL can serve as a plug-in solution to inject explicit modeling of global structure in motion data, we evaluate models for both linear probing and finetuning combining loss terms of existing frameworks with our proposed MorphCL loss term as an unweighted sum. Ablation experiments using different weighting schemes can be found in the Appendix. Results in Table~\ref{tab:small_probingresults_morphimprovement} and~\ref{tab:small_finetuningresults_morphimprovement} show that by making SSL training aware of the morphological similarity through the additional MorphCL loss term, results consistently improve across nearly all methods with linear probing improvements ranging up to as much as 15\% in F1-score.

Notably, REBAR and RelCon, two recent contrastive learning methods that explicitly address positive pair selection in inertial HAR, benefit most substantially from the addition of MorphCL, suggesting that while improved pair retrieval is valuable, it does not fully compensate for the lack of global structural awareness in the embedding space. Similarly, MTL, used to train foundation models for HAR as introduced by~\citet{yuanSelfsupervisedLearningHuman2024}, benefits from the structure-aware contrastive loss by as much as 5\% when pretraining on CAPTURE-24. 
SimCLR benefits from MorphCL on the evaluated subsets (see Section~\ref{sec:ablation}), while no improvement is observed under full-dataset pretraining. We hypothesize that this partly reflects conflicting pair assignments caused by SimCLR's negative pair definition, whereby all in-batch instances are treated as negatives.  
Detailed results of all experiments are provided in the Appendix, including using only MorphCL as loss.

\begin{table}
    \centering
    \scriptsize
    \setlength{\tabcolsep}{3pt}
    \renewcommand{\arraystretch}{1.2}
    \caption{Linear probing macro F1 (\%), averaged over the five downstream datasets. Standard deviations across seeded runs appear below the means. \textbf{Bold}: higher mean per method and encoder/pretraining setting. Dataset-level results are provided in Appendix~\ref{sec:suppexp}.}
    \label{tab:small_probingresults_morphimprovement}
    \begin{tabular}{@{}l*{9}{c}@{}}
        \toprule
        & \multicolumn{3}{c}{\makecell{DeepConvLSTM\\WISDM}}
        & \multicolumn{3}{c}{\makecell{ViT-S\\WISDM}}
        & \multicolumn{3}{c}{\makecell{ViT-S\\CAPTURE-24}} \\
        \cmidrule(lr){2-4} \cmidrule(lr){5-7} \cmidrule(lr){8-10}
        Method & Base & + MorphCL & $\Delta$ & Base & + MorphCL & $\Delta$ & Base & + MorphCL & $\Delta$ \\
        \midrule
        MTL
        & \meanstd{38.31}{0.06} & \meanstd{\mathbf{48.59}}{0.25} & \textcolor{ForestGreen}{$+10.28$} & \meanstd{47.41}{0.22} & \meanstd{\mathbf{50.53}}{0.45} & \textcolor{ForestGreen}{$+3.13$} & \meanstd{47.29}{0.12} & \meanstd{\mathbf{52.12}}{0.33} & \textcolor{ForestGreen}{$+4.83$} \\
        SimCLR
        & \meanstd{46.27}{0.09} & \meanstd{\mathbf{47.85}}{0.26} & \textcolor{ForestGreen}{$+1.58$} & \meanstd{48.78}{0.57} & \meanstd{\mathbf{50.82}}{0.50} & \textcolor{ForestGreen}{$+2.04$} & \meanstd{\mathbf{51.36}}{0.20} & \meanstd{50.97}{0.23} & \textcolor{OrangeRed}{$-0.39$} \\
        Masked AE
        & -- & -- & -- & \meanstd{44.28}{0.87} & \meanstd{\mathbf{50.57}}{0.50} & \textcolor{ForestGreen}{$+6.29$} & \meanstd{47.56}{0.38} & \meanstd{\mathbf{50.05}}{0.37} & \textcolor{ForestGreen}{$+2.49$} \\
        Denoising AE
        & -- & -- & -- & \meanstd{47.73}{0.49} & \meanstd{\mathbf{49.32}}{0.33} & \textcolor{ForestGreen}{$+1.59$} & \meanstd{48.27}{0.23} & \meanstd{\mathbf{51.13}}{0.33} & \textcolor{ForestGreen}{$+2.86$} \\

        REBAR
        & \meanstd{37.81}{0.19} & \meanstd{\mathbf{46.43}}{0.17} & \textcolor{ForestGreen}{$+8.62$} & \meanstd{41.48}{0.68} & \meanstd{\mathbf{49.81}}{0.39} & \textcolor{ForestGreen}{$+8.32$} & \meanstd{37.09}{0.13} & \meanstd{\mathbf{48.27}}{0.49} & \textcolor{ForestGreen}{$+11.18$} \\

        RelCon
        & \meanstd{35.19}{0.36} & \meanstd{\mathbf{45.88}}{0.20} & \textcolor{ForestGreen}{$+10.69$} & \meanstd{34.61}{0.33} & \meanstd{\mathbf{49.96}}{0.40} & \textcolor{ForestGreen}{$+15.36$} & \meanstd{40.52}{0.98} & \meanstd{\mathbf{49.93}}{0.30} & \textcolor{ForestGreen}{$+9.41$} \\
        \bottomrule
    \end{tabular}
\end{table}

\begin{table}
    \centering
    \scriptsize
    \setlength{\tabcolsep}{3pt}
    \renewcommand{\arraystretch}{1.2}
    \caption{Finetuning macro F1 (\%), averaged over the five downstream datasets (mean $\pm$ standard deviation across seeded runs). \textbf{Bold}: higher mean per method and encoder/pretraining setting. Dataset-level results are provided in Appendix~\ref{sec:suppexp}.}
    \label{tab:small_finetuningresults_morphimprovement}
    \begin{tabular}{@{}l*{9}{c}@{}}
        \toprule
        & \multicolumn{3}{c}{\makecell{DeepConvLSTM\\WISDM}}
        & \multicolumn{3}{c}{\makecell{ViT-S\\WISDM}}
        & \multicolumn{3}{c}{\makecell{ViT-S\\CAPTURE-24}} \\
        \cmidrule(lr){2-4} \cmidrule(lr){5-7} \cmidrule(lr){8-10}
        Method & Base & + MorphCL & $\Delta$
               & Base & + MorphCL & $\Delta$
               & Base & + MorphCL & $\Delta$ \\
        \midrule
        MTL
        & \meanstd{54.18}{0.44} & \meanstd{\mathbf{58.07}}{0.22} & \textcolor{ForestGreen}{$+3.89$}
        & \meanstd{57.51}{0.27} & \meanstd{\mathbf{58.23}}{0.16} & \textcolor{ForestGreen}{$+0.71$}
        & \meanstd{60.26}{0.15} & \meanstd{\mathbf{60.68}}{0.11} & \textcolor{ForestGreen}{$+0.42$} \\

        SimCLR
        & \meanstd{57.32}{0.31} & \meanstd{\mathbf{57.57}}{0.28} & \textcolor{ForestGreen}{$+0.24$}
        & \meanstd{55.28}{0.25} & \meanstd{\mathbf{56.25}}{0.19} & \textcolor{ForestGreen}{$+0.97$}
        & \meanstd{58.22}{0.39} & \meanstd{\mathbf{58.72}}{0.19} & \textcolor{ForestGreen}{$+0.51$} \\

        Masked AE
        & -- & -- & --
        & \meanstd{54.36}{0.32} & \meanstd{\mathbf{56.18}}{0.16} & \textcolor{ForestGreen}{$+1.81$}
        & \meanstd{55.61}{0.32} & \meanstd{\mathbf{57.98}}{0.13} & \textcolor{ForestGreen}{$+2.37$} \\

        Denoising AE
        & -- & -- & --
        & \meanstd{53.74}{0.24} & \meanstd{\mathbf{55.27}}{0.13} & \textcolor{ForestGreen}{$+1.53$}
        & \meanstd{55.31}{0.29} & \meanstd{\mathbf{58.35}}{0.25} & \textcolor{ForestGreen}{$+3.04$} \\

        REBAR
        & \meanstd{53.23}{0.29} & \meanstd{\mathbf{57.01}}{0.04} & \textcolor{ForestGreen}{$+3.78$}
        & \meanstd{52.95}{0.21} & \meanstd{\mathbf{54.83}}{0.24} & \textcolor{ForestGreen}{$+1.88$}
        & \meanstd{54.05}{0.13} & \meanstd{\mathbf{58.12}}{0.16} & \textcolor{ForestGreen}{$+4.07$} \\

        RelCon
        & \meanstd{52.82}{0.04} & \meanstd{\mathbf{56.18}}{0.40} & \textcolor{ForestGreen}{$+3.35$}
        & \meanstd{53.43}{0.30} & \meanstd{\mathbf{53.95}}{0.15} & \textcolor{ForestGreen}{$+0.52$}
        & \meanstd{53.50}{0.40} & \meanstd{\mathbf{56.66}}{0.42} & \textcolor{ForestGreen}{$+3.16$} \\
        \bottomrule
    \end{tabular}
\end{table}

\subsection{Ablation Experiments}
\label{sec:ablation}
\textbf{Subset evaluation.} To assess the contributions of data filtering and auxiliary MorphCL training, we train the base SSL objectives on motif-filtered and  morphological subsets with and without MorphCL. The motif-filtered subset consists of all motif windows that were identified to contain a motif as per the motif-discovery. Morphological subset consists of all windows that are part of a morphological cluster and use the same balanced sampler as MorphCL. Table~\ref{tab:subset_wisdm_main} reports results for MTL, SimCLR and Masked AE pretrained on WISDM. Adding MorphCL increases results compared to the based objective. Complete results are provided in Appendix Tables~\ref{tab:subset_wisdm} and~\ref{tab:subset_cap24}.

\begin{table*}
    \centering
    \scriptsize
    \setlength{\tabcolsep}{3pt}
    \renewcommand{\arraystretch}{1.2}
    \caption{Subset pretraining on WISDM (linear probing macro F1 (\%), averaged over five downstream datasets; mean $\pm$ standard deviation across seeded runs). + MorphCL adds the auxiliary MorphCL loss computed on the Morph subset. \textbf{Bold}: higher mean per SSL-subset pair.}
    \label{tab:subset_wisdm_main}
    \begin{tabular}{@{}l*{9}{c}@{}}
        \toprule
        & \multicolumn{3}{c}{MTL}
        & \multicolumn{3}{c}{SimCLR}
        & \multicolumn{3}{c}{Masked AE} \\
        \cmidrule(lr){2-4}
        \cmidrule(lr){5-7}
        \cmidrule(lr){8-10}
        Subset
        & Base & + MorphCL & $\Delta$
        & Base & + MorphCL & $\Delta$
        & Base & + MorphCL & $\Delta$ \\
        \midrule
        Full
        & \meanstd{47.41}{0.22}
        & \meanstd{\mathbf{50.53}}{0.45}
        & \textcolor{ForestGreen}{$+3.13$}
        & \meanstd{48.78}{0.57}
        & \meanstd{\mathbf{50.82}}{0.50}
        & \textcolor{ForestGreen}{$+2.04$}
        & \meanstd{44.28}{0.87}
        & \meanstd{\mathbf{50.57}}{0.50}
        & \textcolor{ForestGreen}{$+6.29$} \\

        Motif
        & \meanstd{46.13}{0.36}
        & \meanstd{\mathbf{49.70}}{0.79}
        & \textcolor{ForestGreen}{$+3.57$}
        & \meanstd{47.75}{0.62}
        & \meanstd{\mathbf{50.53}}{0.69}
        & \textcolor{ForestGreen}{$+2.78$}
        & \meanstd{43.56}{0.66}
        & \meanstd{\mathbf{50.06}}{0.47}
        & \textcolor{ForestGreen}{$+6.50$} \\

        Morph
        & \meanstd{46.86}{0.36}
        & \meanstd{\mathbf{49.42}}{0.49}
        & \textcolor{ForestGreen}{$+2.56$}
        & \meanstd{47.88}{0.62}
        & \meanstd{\mathbf{49.80}}{0.50}
        & \textcolor{ForestGreen}{$+1.92$}
        & \meanstd{42.42}{0.87}
        & \meanstd{\mathbf{50.12}}{0.88}
        & \textcolor{ForestGreen}{$+7.70$} \\
        \bottomrule
    \end{tabular}
\end{table*}

\textbf{Hyperparameter sensitivity.} We independently vary the motif-discovery parameters ($\sigma$, $r$) and clustering parameters ($\rho$, support) on WISDM and test linear probing results on our five chosen downstream tasks (see Figure~\ref{fig:ablation_sensitivity}; see Appendix~\ref{sec:suppexp} for full results and additional analysis). While our chosen default configurations show highest results, our implementation remains stable across various setups. At $\rho=400$, all samples are classified as noise, preventing downstream evaluation.

\begin{figure*}
    \centering
    \includegraphics[width=0.9\textwidth]{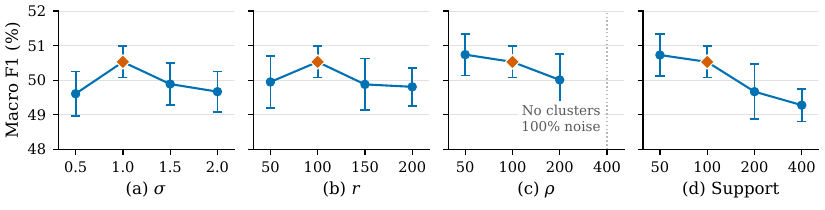}
    \caption{Sensitivity of MTL + MorphCL on WISDM. Mean linear-probing macro F1 (\%) over five downstream datasets. Diamonds mark defaults: $\sigma=1$, $r=\rho=\mathrm{support}=100$.}
    \label{fig:ablation_sensitivity}
\end{figure*}

\subsection{Qualitative Results}

\begin{figure}
    \centering
    \includegraphics[width=1.0\linewidth]{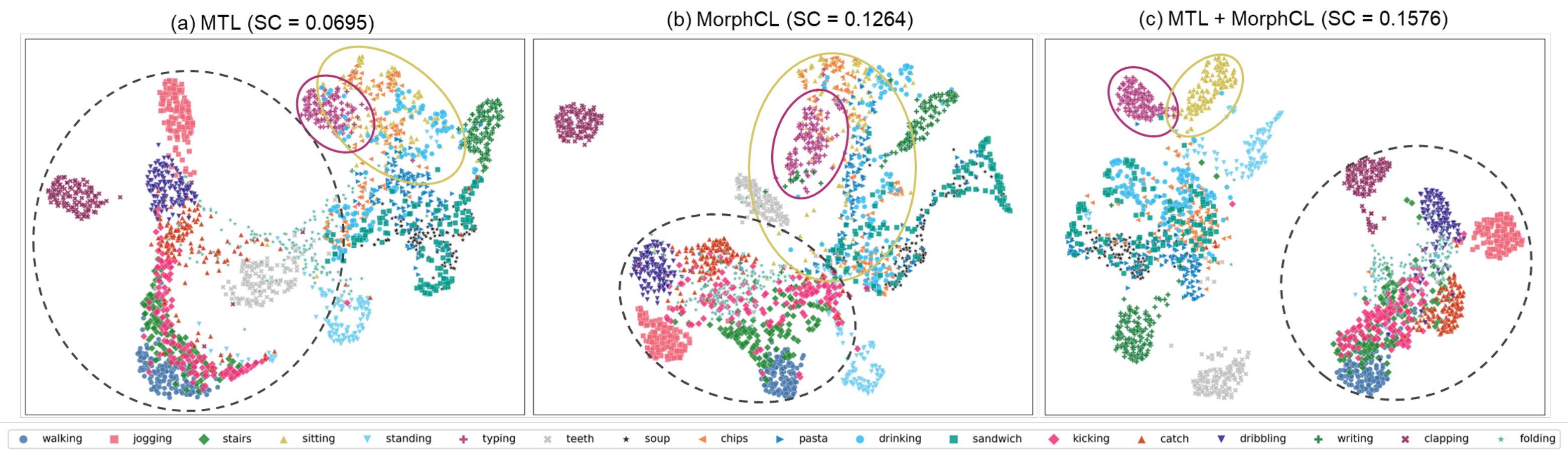}
    \caption{UMAP~\citep{mcinnesUMAPUniformManifold2020} embeddings of a WISDM participant from ViT-S models pretrained on CAPTURE-24 using (i) MTL, (ii) MorphCL and (iii) MTL + MorphCL. SC: silhouette coefficient ($[-1,1]$; higher is better). On unseen data, MTL yields entangled clusters, while MorphCL captures global structure. Their combination improves cluster compactness and separation (higher SC), distinguishing sitting from typing (yellow/purple circles) while preserving coherent manifolds among vigorous activities (dashed circle).}
    \label{fig:umap_wisdm}
\end{figure}

To better illustrate what MorphCL learns, Figure~\ref{fig:umap_wisdm} shows how representations pretrained on CAPTURE-24 transfer to WISDM for a sample participant, visualized against activity annotations unseen during training. In (a) MTL, some coarse grouping exists, but the embedding space remains largely entangled: kinematically similar activities, e.g., typing vs. sitting (yellow and purple circles), overlap substantially. In (b) MorphCL, the global structure becomes more coherent, with samples organized into smoother manifolds that reflect morphological consistency, e.g. vigorous activities (dashed circle) separate more clearly from seated activities, yet class boundaries remain diffuse, and activities sharing similar motion patterns occupy contiguous, poorly separated regions. The combined objective ((c) MorphCL + MTL) yields both structure and separability: clusters are compact and well-delineated, with clearer boundaries even between kinematically similar actions. 

This qualitative trend is further supported quantitatively by silhouette coefficients~\citep{rousseeuwSilhouettesGraphicalAid1987} provided with the plot and computed across all downstream datasets (full results in appendix), which consistently increase with MorphCL. Linear probing results confirm this further (see Appendix~\ref{sec:extendedqualatative} for confusion matrices) showing that residual misclassifications concentrate among semantically related activities rather than being scattered indiscriminately.

\subsection{Comparison to Existing Accelerometer Foundation Models}

While several efforts have been made in other communities advancing towards foundational models which can be used to extract discriminative features, open-sourced inertial-based foundation models remain sparse. Most existing approaches are closed-source due to data privacy constraints~\citep{xuRelConRelativeContrastive2025, parkSlotFMMotionFoundation2025, narayanswamy2025scaling, zhangSensorLMLearningLanguage2025} and/or are applied using window sizes of several minutes~\citep{logacjovSelfPABLargescalePretraining2024}, making them not suited for accelerometer-based in-situ activity recognition. To our knowledge, the only publicly available model for this setting is provided by \citet{yuanSelfsupervisedLearningHuman2024}, who release ResNet-18 weights pretrained on the  UK Biobank dataset under controlled access (100k participants, one-week wrist accelerometer recordings) and public CAPTURE-24 dataset using the MTL framework. 

Table~\ref{tab:foundation_comparison} compares model weights provided by~\citet{yuanSelfsupervisedLearningHuman2024} with a selection of different sized ViT (as used by~\citet{hoddesScalingLawsWearable2025}) and a ResNet-18 backbone pretrained using MTL combined with MorphCL on CAPTURE-24. 
Because~\citet{yuanSelfsupervisedLearningHuman2024} use larger input windows (10s vs. our 2.56s), we average embeddings over four consecutive windows (10.24s) for comparability. 
Despite using a smaller model (ViT-S), we achieve comparable probing performance to the Biobank model and surpass it when using the same ResNet-18 backbone, despite training on a dataset $4600\times$ smaller. This result indicates that incorporating global structural information can partially compensate for reduced pretraining scale by using the available data more efficiently.

\begin{table*}
\centering
\tiny
\setlength{\tabcolsep}{3pt}
\caption{Comparison with pretrained models from~\citet{yuanSelfsupervisedLearningHuman2024}. Macro F1 (\%; mean $\pm$ standard deviation across seeded runs) is reported for linear probing and finetuning.}
\label{tab:foundation_comparison}
\resizebox{\textwidth}{!}{%
\begin{tabular}{l*{14}{c}}
\toprule
& \makecell[c]{Hours\\of Data}
& \makecell[c]{Model\\Size}
& \multicolumn{2}{c}{WEAR}
& \multicolumn{2}{c}{Wetlab}
& \multicolumn{2}{c}{PAMAP2}
& \multicolumn{2}{c}{Hangtime}
& \multicolumn{2}{c}{RWHAR}
& \multicolumn{2}{c}{\textit{Avg.}} \\
\cmidrule(lr){4-5} \cmidrule(lr){6-7}
\cmidrule(lr){8-9} \cmidrule(lr){10-11}
\cmidrule(lr){12-13} \cmidrule(lr){14-15}
& & & LProbe & Fine
& LProbe & Fine
& LProbe & Fine
& LProbe & Fine
& LProbe & Fine
& LProbe & Fine \\
\midrule
\multicolumn{15}{l}{\citet{yuanSelfsupervisedLearningHuman2024} - MTL} \\
\midrule
\makecell[l]{ResNet18} & 3.7K & 10.4M
& \meanstd{33.77}{0.24} & \meanstd{77.68}{0.56}
& \meanstd{18.53}{0.35} & \meanstd{41.38}{0.37}
& \meanstd{54.15}{0.68} & \meanstd{75.01}{2.76}
& \meanstd{24.54}{0.32} & \meanstd{29.75}{0.14}
& \meanstd{50.22}{0.61} & \meanstd{79.21}{0.46}
& \meanstd{36.24}{0.18} & \meanstd{60.61}{0.60} \\

\rowcolor{gray!15}
\makecell[l]{ResNet18} & 16.8M
& 10.4M
& \meanstd{71.45}{0.16} & \meanstd{83.86}{0.41}
& \meanstd{35.00}{0.49} & \meanstd{48.13}{0.42}
& \meanstd{77.06}{0.62} & \meanstd{84.00}{1.38}
& \meanstd{29.80}{0.22} & \meanstd{29.68}{0.07}
& \meanstd{76.70}{0.60} & \meanstd{86.30}{0.63}
& \meanstd{58.00}{0.05} & \meanstd{66.39}{0.39} \\
\midrule
\multicolumn{15}{l}{\textbf{Ours - MTL + MorphCL}} \\
\midrule
\makecell[l]{ResNet18} & 3.7K & 10.4M
& \meanstd{72.40}{0.21} & \meanstd{80.61}{0.14}
& \meanstd{39.98}{0.61} & \meanstd{47.02}{0.54}
& \meanstd{75.15}{1.07} & \meanstd{75.80}{1.71}
& \meanstd{29.64}{0.07} & \meanstd{28.73}{0.09}
& \meanstd{76.03}{1.05} & \meanstd{80.74}{1.59}
& \meanstd{58.64}{0.31} & \meanstd{62.58}{0.27} \\

\makecell[l]{ViT-S} & 3.7K & 1M
& \meanstd{65.82}{0.48} & \meanstd{75.85}{0.69}
& \meanstd{31.68}{0.19} & \meanstd{44.92}{0.23}
& \meanstd{75.28}{0.48} & \meanstd{79.88}{1.40}
& \meanstd{29.57}{0.35} & \meanstd{29.90}{0.20}
& \meanstd{75.46}{0.28} & \meanstd{77.10}{0.47}
& \meanstd{55.56}{0.16} & \meanstd{61.53}{0.31} \\

\makecell[l]{ViT-M} & 3.7K & 7.9M
& \meanstd{63.94}{0.26} & \meanstd{79.61}{0.26}
& \meanstd{30.66}{0.85} & \meanstd{47.47}{0.82}
& \meanstd{72.03}{0.55} & \meanstd{81.71}{1.19}
& \meanstd{30.20}{0.15} & \meanstd{29.89}{0.36}
& \meanstd{77.80}{0.21} & \meanstd{82.42}{0.85}
& \meanstd{54.92}{0.18} & \meanstd{64.22}{0.52} \\

\makecell[l]{ViT-L} & 3.7K & 63.7M
& \meanstd{63.42}{0.46} & \meanstd{79.35}{0.38}
& \meanstd{26.25}{0.76} & \meanstd{46.37}{0.59}
& \meanstd{73.17}{0.32} & \meanstd{79.40}{2.76}
& \meanstd{29.65}{0.36} & \meanstd{17.96}{2.09}
& \meanstd{71.89}{0.75} & \meanstd{78.40}{0.51}
& \meanstd{52.87}{0.07} & \meanstd{60.30}{0.15} \\
\bottomrule
\end{tabular}%
}
\end{table*}

\subsection{Limitations}
\label{ch:limi}
To evaluate our proposed MorphCL framework, we use CAPTURE-24, the largest publicly available dataset for this setting, but its scale remains modest relative to private pretraining corpora. This limitation is most apparent for larger ViT variants: while finetuning improves for ViT-M, linear probing does not.
While our motif discovery and clustering hyperparameters transfer across CAPTURE-24 and WISDM, we have not studied scalability to substantially larger datasets. We leave this to future work, as we expect MorphCL to support more stable scaling on larger in-the-wild corpora. 
Regardless of the size of pretraining data, WISDM results indicate that already a few instances of activity executions per person can be sufficient for learning underlying morphological structures.
Further, while we only test our MorphCL framework on accelerometer data, matrix profiling frameworks such as~\citet{yehMatrixProfileVI2017} can be analogously applied to magnetometer and gyroscope data.

\section{Conclusion}
This work introduced MorphCL, a group-based contrastive objective that uses two well-established pillars of motion analysis to inject domain knowledge into self-supervised pretraining for inertial data. Compared to existing SSL frameworks our approach grounds its notion of similarity in domain-informed feature descriptors and moves beyond instance-level and random comparisons, leveraging latent morphological structures present in inertial pretraining datasets.
Concretely, we combine motif discovery, a technique with a long history in HAR for identifying the building blocks of complex motion, with domain-specific feature clustering to partition two large-scale pretraining corpora, WISDM and CAPTURE-24, into morphological groups, and use these groups to define an auxiliary contrastive loss that complements rather than replaces existing SSL frameworks. Downstream experiments demonstrate that incorporating MorphCL as an additional loss head improves representation quality of existing SSL frameworks, with gains of up to 15\% in F1-score. A comparison to the accelerometer foundation model trained on the Biobank dataset~\citep{yuanSelfsupervisedLearningHuman2024} shows that our method is capable of beating probing results while being trained on $4600\times$ less data. 

\bibliography{ICLR/main}
\bibliographystyle{iclr2027_conference}

\newpage

\appendix

\section{Technical Appendices and Supplementary Material}
\label{ch:technicalapp}

\subsection{Motif-discovery \& Morphological Clustering}
In order to discover repeated patterns within the WISDM and CAPTURE-24 dataset we make use of the motif-discovery tool STUMPY~\citep{law2019stumpy} which implements \emph{multi-dimensional matrix profiling} as suggested by~\citet{yehMatrixProfileVI2017}. STUMPY efficiently computes the matrix profile, which stores for any subsequence its distance to the most similar non-trivial subsequence, based on pairwise, z-score normalized Euclidean distances. This allows determining the $r$ closest reoccurrences of a motif $m_i$ which do not exceed a pre-defined motif distance threshold $\sigma$. As datasets such as the CAPTURE-24 dataset are highly unbalanced in the occurrences of their repeated patters, with e.g. flat line motifs dominating large stretches of the recorded timeseries, we mask out any additional subsequence whose matrix profile distance also lies within $\sigma$ from the selected motif by setting their matrix profile values to infinity before extracting the next most prominent motif in the input timeseries. In order to scale our motif discovery to large inertial datasets, we use STUMPY’s Dask-backed implementation, which enables the matrix profile to be computed in a chunked and parallelized fashion across multiple CPU cores or nodes. For both the WISDM and CAPTURE-24 dataset we extract $r=100$ occurrences of the top $m=100$ most prominent motifs per chunk, with $\sigma$ being set to $1.0$. We set the repeated subsequence of each motif to be exactly $0.2$ seconds in length extracting a motif window of $W=2.56$ seconds and context window $C=7.68$ seconds around it. Note while sequentially extracting windows based on the extracted motif centers we ignore windows which overlap more than 50\% with already extracted windows. In case of the WISDM dataset we set the chunks to be full-participant recording timelines (as this roughly equates to one hour per participant), while for CAPTURE-24 we set it to roughly one hour of data.

Once all extracted motif windows of each participant and data chunk are collected, we extract features that describe the morphology of each window. We make use of the features as reported by~\citet{vanderdoncktNotSleepTraditional2023} which have shown to reapply these field-tested features to various application scenarios from sleep scoring~\citep{vanderdoncktNotSleepTraditional2023}, mode of transport recognition~\citep{vanderdoncktMagnitudeRotationInvariant2024}, as well as sports activity recognition~\citep{vanderdoncktLeftRightSwappingUpperLower2024}.
Features are grouped into three complementary categories:
\begin{enumerate}
    \item \textit{Distribution features} capture the overall amplitude characteristics and statistical shape of the signal within a window. These include \textit{minimum}, \textit{maximum}, \textit{peak-to-peak} amplitude, \textit{interquartile range}, \textit{standard deviation}, \textit{skewness}, and \textit{kurtosis}.
    \item \textit{Temporal and dynamical features} characterize the short-term evolution and structural complexity. \textit{Hjorth mobility} and \textit{complexity} quantify signal smoothness and oscillatory behavior, while \textit{mean crossing rate} and \textit{Petrosian fractal dimension} capture temporal variability and irregularity in the signal trajectory.
    \item \textit{Spectral and complexity features} summarize the distribution and organization of signal energy. \textit{Spectral entropy}, computed from the Fourier spectrum, and \textit{differential entropy} quantify signal uncertainty and spread, while \textit{Katz fractal dimension} provides a complementary geometric measure of complexity.
\end{enumerate}
The resulting feature vectors describing each window are basis for the morphological clustering. Undefined or non-finite feature values arising from degenerate windows (e.g., flat-line segments) are handled per feature as follows: \textit{skewness} is set to $0$, \textit{kurtosis} to $3$ (the Gaussian reference value), \textit{Hjorth mobility} and \textit{complexity} to $0$, and \textit{differential entropy} to a lower bound defined as the minimum value of said feature among all computed values minus $\mu=0.001$. As reported in the main paper we use HDBSCAN as our clustering algorithm of choice. Being an extension of DBSCAN ~\citep{esterDensitybasedAlgorithmDiscovering1996}, HDBSCAN identifies clusters from the density structure of the feature vectors and treats unassigned windows as noise. We use Euclidean distance and leaf cluster selection, with $\rho=100$ (min\_cluster\_count) and support$=100$ (min\_samples) in the default configuration. Sensitivity to these parameters is examined in Appendix Tables~\ref{tab:changing_rho} and~\ref{tab:changing_support}. With computed domain-specific features possessing widely varying natural scales, we apply z-score normalization on the computed feature vectors. Table~\ref{tab:pretrainsizes} reports sizes of the created pretraining corpora. 

\begin{table}[!ht]
    \centering
    \scriptsize
    \caption{Size comparison of the different pretraining corpora. We compare the unfiltered version of the WISDM and CAPTURE-24 dataset, created by evenly splitting the dataset in windows of $2.56s$ without overlap, with our proposed motif-filtered variant as described in Chapter~\ref{sec:motiffil} and the morphological-clustered variant as described in Chapter~\ref{sec:morphcl}. Values are provided as hours of data.}
    \begin{tabular}{ccccccc}
        \multirow{2}{*}{Pretraining Corpus} & \multicolumn{2}{c}{Unfiltered} & \multicolumn{2}{c}{Motif-filtered} & \multicolumn{2}{c}{Morph-filtered} \\ 
                   & Train  & Test  & Train & Test  & Train & Test \\ \midrule
        WISDM      & 35.5   & 9.7   & 15.2  & 4.4   & 3.2   & 0.8  \\
        CAPTURE-24 & 3004.6 & 787.4 & 783.5 & 197.7 & 47.3  & 39.0 \\
    \end{tabular}
    \label{tab:pretrainsizes}
\end{table}

\subsection{Pretraining} 

During pretraining DeepConvLSTM and ViT backbone architectures were trained using an Adam optimizer with the AMSGrad variant enabled and the initial learning rate set to $5e^{-4}$ for $100$ epochs. Details of each backbone architecture can be found in their original publications~\citep{ordonezDeepConvolutionalLSTM2016, hoddesScalingLawsWearable2025}.
To stabilize training a linear warm-up schedule is applied for the first $10$ epochs followed by a learning rate scheduler that monitors the validation loss and reduces the learning rate by a factor of $0.5$ if it did not improve for $5$ consecutive epochs. The minimum learning rate is clipped at $1e^{-6}$ and a relative improvement threshold of $1e^{-4}$ is used. 
During all pretraining experiments we employ a batch size of $512$ windows. 
Note that the REBAR~\citep{xuREBARRetrievalbasedReconstruction2024} and RelCon~\citep{xuRelConRelativeContrastive2025} SSL framework follow a different batch definition, as their losses are calculated based on one randomly drawn anchor window as well as $20$ candidate windows of each participant during an epoch. To make batch sizes used during REBAR and RelCon thus comparable to other frameworks mentioned in this paper, we draw from $24$ participants to create a REBAR batch, with each participant appearing exactly once per epoch. As this makes epochs restricted in size by the number of participants in the pretrain dataset, we increase the total number of epochs to $1000$ for both REBAR and RelCon, making the total number of windows drawn per pretraining experiment comparable to that of other pretraining framework such as MTL.
In case of our MorphCL framework, to make each batch-wise gradient update an assessment of how well our feature encoder is capable of mapping to a latent space that respects morphological structures of our pretrain corpus, we design batches used during training of the MorphCL head in a way that each morphological group label $y_i$ is represented in a balanced fashion. As groups, however, vary in size, we allow windows to occur multiple times across batches allowing for each window of each morphological group to appear at least once during training. Note that our proposed MorphCL framework uses the same model head structure as SimCLR~\citep{tangExploringContrastiveLearning2021}, differing only in the set of possible augmentations (as mentioned in the main paper) and is thus comparable in complexity to said method. During all experiments we set the temperature value $\tau$ of our MorphCL head to $0.1$.

In total we investigated six SSL frameworks:

\textbf{Multi-task learning (MTL)~\citep{saeedMultitaskSelfSupervisedLearning2019, yuanSelfsupervisedLearningHuman2024}:} binary prediction of whether a set of pre-defined augmentations have been applied to a accelerometer signal or not. As proposed by the original authors the transformations are: flipping the data in the time-dimension, scaling values by a random factor, random permutation of sensor axes and warping the input data. Same as the original publication, we make each augmentation prediction head a 2-layered MLP with a hidden dimension of 128 and ReLU activation.

\textbf{SimCLR~\citep{tangExploringContrastiveLearning2021}:} Similar to image data, the SimCLR framework for accelerometer signals is applied by creating positive pairs as two augmented views of an input signal and negative pairs as all other instances within a batch. As suggested by~\citet{tangExploringContrastiveLearning2021} we chose a total of eight sensor-specific augmentations as our set of possible augmentations. Same as~\citet{tangExploringContrastiveLearning2021} we use a 2-layered MLP with a 512 hidden dimension, ReLU activation, temperature $\tau=0.1$ and 128 output dimension.

\textbf{Denoising Autoencoder~\citep{haresamudramRoleFeaturesHuman2019}:} consists of using an encoder-decoder structure to recover an original input signal. We additionally randomly augment input signals, choosing from the same set of augmentations used in SimCLR, with the model needing to recover the unaugmented view. We reimplement the decoder as detailed in~\citet{hoddesScalingLawsWearable2025}, using $2$ decoding layers, $4$ attention heads, $0.1$ dropout, GELU activation and setting the decoder dimension to half the input dimension and feedforward dimension to twice the decoder dimension. 

\textbf{Masked Autoencoder~\citep{haresamudramMaskedReconstructionBased2020, hoddesScalingLawsWearable2025}:} aims to learn useful latent features by having models reconstruct masked parts of an input signal through patterns extracted from surrounding unmasked parts of the input signal. Same as the denoising decoder, we reimplement the decoder as detailed in~\citet{hoddesScalingLawsWearable2025}, applying also a token-based masking ratio of $0.7$. 

\textbf{REBAR~\citep{xuREBARRetrievalbasedReconstruction2024}:} Extending the idea of SimCLR~\citep{xuREBARRetrievalbasedReconstruction2024}, introduced REBAR, which provides a learned metric to determine positive pairs within a batch. Similar to our approach, their work is based on the assumption that windows which are capable of reconstructing masked versions of each other well share inherent patterns. That way, the framework provides a way to define positive pairs across signals as well as participants. Note that negative pairs are defined the same way as in SimCLR. We reuse the implementation of REBAR from the authors repository, setting the embedding dimension to 256 and double receptive field parameter to 1. To calculate the REBAR loss we use $\tau=0.1$ and $\alpha=0.0$ as done in the original publication.

\textbf{RelCon~\citep{xuRelConRelativeContrastive2025}:} In addition to small improvements to the training of the similarity metric and loss calculation, RelCon softens the negative pair definition of REBAR by replacing it with a function that dynamically adjusts the negative pairs based on the current positive pairing. This allows RelCon to dynamically learn a ranking among signals in a batch rather than relying on hard positive and negative pair decisions. We reuse the implementation of RelCon from the authors repository. To train the involved REBAR metric we reuse settings from the REBAR repository, i.e. setting the embedding dimension to 256 and double receptive field parameter to 1. To calculate the RelCon loss we use $\tau=0.1$ as done in the original publication.

\subsection{Downstream Experiments}
During all downstream experiments, pretraining heads, such as our proposed MorphCL head, were removed from the backbone and replaced by a single linear classification layer, which maps the output dimension of the last backbone layer directly to the number of classes present in the downstream dataset. In case of linear probing experiments, we freeze the weights of the backbone allowing only the linear classification layer to be trained, while for finetuning experiments, we additionally allow all layers of the backbone to be adjusted. 
We use an Adam optimizer with the initial learning rate being $1e^{-1}$ for the linear classification head and $1e^{-4}$ for finetuning the pretrained backbone networks. In line with best practices we employ a Leave-One-Subject-Out cross-validation, where each participant in the dataset becomes the test set exactly once while remaining participants are used for training. Each dataset is segmented using a sliding window of of length $2.56$ seconds with an overlap of $50\%$. Batches consist of $128$ randomly-shuffled windows and are z-score normalized per-axis based on the training data. Each split is trained for $30$ epochs with learning rates being decreased by a factor of $0.5$ if the validation loss, computed on a excluded participant in the train dataset, does not improve over $5$ epochs. Loss calculation is based on a class-weighted cross-entropy loss to accommodate the unbalancedness of the tested downstream datasets. For comparison we also provide fully-supervised training results on each dataset and backbone using a unified initial learning rate for backbone and classifier ($1e^{-3}$). 

\subsection{Licensing \& Runtime Environment}
\label{sec:licensing}
All datasets were downloaded from their respective code repositories and websites as of submitting this paper. All datasets are hosted under either the CC-BY 4.0 (CAPTURE-24, Hang-Time, PAMAP2, RWHAR, Wetlab, WISDM) or the CC BY-NC-SA 4.0 license (WEAR). All experiments were performed on a single-GPU (NVIDIA Tesla V100) with 2 AMD EPYC 7452 CPUs and 256 GB RAM. Larger ViT variants were trained on larger single GPU (NVIDIA A6000) with two AMD EPYC 7413 and 1008 GB RAM. Note that the footprint and computational complexity of each model is dependent on the chosen SSL framework. Motif discovery was performed on one hour chunks of data at a time and took around 170 seconds per chunk performed on a single, standard CPU (Intel Core Ultra 9 285K).

\section{Supplementary Experiments}
\label{sec:suppexp}
\subsection{Detailed Main Results}
Table~\ref{tab:results_method} reports detailed results of the linear probing and finetuning experiments across the five downstream datasets comparing as a standalone framework MorphCL with the tested instance-based learning schemes. Table~\ref{tab:probingresults_morphimprovement} and Table~\ref{tab:finetuningresults_morphimprovement} provide detailed results of the relative linear probing and finetuning improvements achieved when combining MorphCL with the instance-based learning schemes. Note that results are provided as macro F1-scores averaged across the individual LOSO splits. Each experiment was repeated three times using a set of three different random seeds. The last column of each table further provides the average across all downstream datasets. To characterize variability across the three random seeds, we report standard deviations of all experiments.

\begin{table*}[!ht]
\centering
\tiny
\caption{Downstream result comparison of the proposed morphological learning with instance-based self-supervised learning methods. Results are reported for five downstream benchmark datasets. Pretraining was performed on both the WISDM and CAPTURE-24 dataset using different encoder backbones. Fully-supervised learning refers to training the backbone without prior pretraining directly on the downstream dataset. Macro F1 is reported in \% as mean $\pm$ standard deviation across seeded runs. Shaded rows denote standalone MorphCL.}
\label{tab:results_method}
\setlength{\tabcolsep}{3pt}
\renewcommand{\arraystretch}{1.2}
\resizebox{\textwidth}{!}{%
\begin{tabular}{l*{12}{c}}
\toprule
\multirow{2}{*}{}
& \multicolumn{2}{c}{WEAR}
& \multicolumn{2}{c}{Wetlab}
& \multicolumn{2}{c}{PAMAP2}
& \multicolumn{2}{c}{Hangtime}
& \multicolumn{2}{c}{RWHAR} 
& \multicolumn{2}{c}{\textit{Avg.}}\\
\cmidrule(lr){2-3} \cmidrule(lr){4-5} \cmidrule(lr){6-7} \cmidrule(lr){8-9} \cmidrule(lr){10-11} \cmidrule(lr){12-13}
& LProbe & Fine
& LProbe & Fine
& LProbe & Fine
& LProbe & Fine
& LProbe & Fine
& LProbe & Fine \\
\midrule
\multicolumn{13}{l}{\textbf{Encoder:} DeepConvLSTM} \\
\textit{Fully-supervised}  & \multicolumn{2}{c}{\textit{\meanstd{\mathit{61.42}}{0.63}}} & \multicolumn{2}{c}{\textit{\meanstd{\mathit{29.34}}{0.38}}} & \multicolumn{2}{c}{\textit{\meanstd{\mathit{65.46}}{0.93}}} & \multicolumn{2}{c}{\textit{\meanstd{\mathit{52.04}}{0.31}}} & \multicolumn{2}{c}{\textit{\meanstd{\mathit{63.89}}{0.70}}} & \multicolumn{2}{c}{\textit{\meanstd{\mathit{54.43}}{0.17}}} \\
\midrule
\multicolumn{13}{l}{\textbf{Pretraining:} WISDM} \\
\textit{MTL}              & \meanstd{37.72}{0.34} & \meanstd{57.86}{0.53} & \meanstd{17.63}{0.07} & \meanstd{27.99}{0.19} & \meanstd{54.84}{0.18} & \meanstd{69.69}{0.16} & \meanstd{27.03}{0.56} & \meanstd{48.59}{0.23} & \meanstd{54.31}{0.44} & \meanstd{66.77}{1.46} & \meanstd{38.31}{0.06} & \meanstd{54.18}{0.44} \\
\textit{SimCLR}           & \meanstd{46.74}{0.11} & \meanstd{60.66}{0.25} & \meanstd{21.21}{0.41} & \meanstd{32.58}{0.25} & \meanstd{61.47}{0.22} & \meanstd{73.64}{0.80} & \meanstd{38.64}{0.36} & \meanstd{52.40}{0.12} & \meanstd{63.29}{0.45} & \meanstd{67.35}{0.89} & \meanstd{46.27}{0.09} & \meanstd{57.32}{0.31} \\
\textit{REBAR}            & \meanstd{33.64}{0.26} & \meanstd{56.64}{0.20} & \meanstd{18.16}{0.05} & \meanstd{27.76}{0.08} & \meanstd{53.86}{0.28} & \meanstd{67.06}{0.64} & \meanstd{24.25}{0.22} & \meanstd{50.89}{0.39} & \meanstd{59.15}{0.77} & \meanstd{63.81}{0.75} & \meanstd{37.81}{0.19} & \meanstd{53.23}{0.29} \\
\textit{RelCon}           & \meanstd{30.80}{0.11} & \meanstd{56.43}{0.15} & \meanstd{16.19}{0.11} & \meanstd{27.22}{0.54} & \meanstd{48.90}{0.63} & \meanstd{65.24}{1.01} & \meanstd{25.39}{0.24} & \meanstd{50.87}{0.23} & \meanstd{54.66}{1.23} & \meanstd{64.37}{1.24} & \meanstd{35.19}{0.36} & \meanstd{52.82}{0.04} \\
\addlinespace[1pt]
\rowcolor{gray!15}
\textbf{MorphCL}                       & \meanstd{43.99}{0.42} & \meanstd{60.00}{0.37} & \meanstd{24.24}{0.11} & \meanstd{31.62}{0.41} & \meanstd{60.85}{0.23} & \meanstd{70.27}{1.37} & \meanstd{32.84}{0.52} & \meanstd{52.16}{0.50} & \meanstd{65.20}{1.10} & \meanstd{69.47}{1.46} & \meanstd{45.42}{0.13} & \meanstd{56.71}{0.17} \\
\midrule
\multicolumn{13}{l}{\textbf{Encoder:} ViT-S} \\ 
\textit{Fully-supervised}  & \multicolumn{2}{c}{\textit{\meanstd{\mathit{63.76}}{0.15}}} & \multicolumn{2}{c}{\textit{\meanstd{\mathit{25.20}}{0.51}}} & \multicolumn{2}{c}{\textit{\meanstd{\mathit{66.44}}{0.46}}} & \multicolumn{2}{c}{\textit{\meanstd{\mathit{34.10}}{1.80}}} & \multicolumn{2}{c}{\textit{\meanstd{\mathit{63.90}}{0.47}}} & \multicolumn{2}{c}{\textit{\meanstd{\mathit{50.68}}{0.25}}} \\
\midrule
\multicolumn{13}{l}{\textbf{Pretraining:} WISDM} \\ 
\textit{MTL}              & \meanstd{50.05}{0.59} & \meanstd{66.02}{0.33} & \meanstd{22.39}{0.24} & \meanstd{33.25}{0.30} & \meanstd{63.84}{0.42} & \meanstd{70.05}{0.70} & \meanstd{33.71}{0.98} & \meanstd{49.67}{0.15} & \meanstd{67.03}{0.80} & \meanstd{68.58}{0.45} & \meanstd{47.41}{0.22} & \meanstd{57.51}{0.27} \\
\textit{SimCLR}           & \meanstd{53.08}{0.62} & \meanstd{63.12}{0.41} & \meanstd{25.94}{0.41} & \meanstd{32.71}{0.24} & \meanstd{64.00}{1.25} & \meanstd{67.35}{0.87} & \meanstd{35.73}{0.44} & \meanstd{47.96}{0.30} & \meanstd{65.14}{1.44} & \meanstd{65.25}{0.85} & \meanstd{48.78}{0.57} & \meanstd{55.28}{0.25} \\
\textit{Masked AE}        & \meanstd{46.58}{0.80} & \meanstd{60.57}{0.07} & \meanstd{17.43}{1.04} & \meanstd{31.09}{0.08} & \meanstd{57.86}{0.65} & \meanstd{66.59}{1.82} & \meanstd{37.19}{1.60} & \meanstd{47.65}{0.14} & \meanstd{62.32}{1.03} & \meanstd{65.92}{0.16} & \meanstd{44.28}{0.87} & \meanstd{54.36}{0.32} \\
\textit{Denoising AE}     & \meanstd{49.70}{0.59} & \meanstd{60.51}{0.08} & \meanstd{20.92}{0.93} & \meanstd{31.25}{0.16} & \meanstd{62.53}{0.25} & \meanstd{65.63}{1.36} & \meanstd{38.35}{0.38} & \meanstd{47.21}{0.24} & \meanstd{67.15}{1.08} & \meanstd{64.11}{0.64} & \meanstd{47.73}{0.49} & \meanstd{53.74}{0.24} \\
\textit{REBAR}            & \meanstd{42.59}{0.68} & \meanstd{58.84}{0.06} & \meanstd{16.80}{0.40} & \meanstd{30.13}{0.30} & \meanstd{59.01}{0.38} & \meanstd{64.95}{1.14} & \meanstd{29.12}{0.38} & \meanstd{46.70}{0.33} & \meanstd{59.89}{2.57} & \meanstd{64.11}{0.68} & \meanstd{41.48}{0.68} & \meanstd{52.95}{0.21} \\
\textit{RelCon}           & \meanstd{37.53}{0.63} & \meanstd{59.42}{0.16} & \meanstd{13.70}{0.61} & \meanstd{30.20}{0.11} & \meanstd{43.71}{1.19} & \meanstd{64.54}{0.95} & \meanstd{25.88}{0.26} & \meanstd{47.30}{0.01} & \meanstd{52.21}{1.18} & \meanstd{65.69}{0.87} & \meanstd{34.61}{0.33} & \meanstd{53.43}{0.30} \\
\addlinespace[1pt]
\rowcolor{gray!15}
\textbf{MorphCL}                       & \meanstd{52.38}{0.09} & \meanstd{61.67}{0.24} & \meanstd{26.45}{0.52} & \meanstd{32.31}{0.15} & \meanstd{66.21}{0.85} & \meanstd{67.52}{1.06} & \meanstd{34.86}{0.81} & \meanstd{47.55}{0.25} & \meanstd{67.92}{0.35} & \meanstd{64.19}{0.70} & \meanstd{49.56}{0.42} & \meanstd{54.65}{0.30} \\
\midrule
\multicolumn{13}{l}{\textbf{Pretraining:} CAPTURE-24} \\
\textit{MTL}               & \meanstd{49.34}{0.74} & \meanstd{69.07}{0.12} & \meanstd{22.44}{0.20} & \meanstd{35.79}{0.04} & \meanstd{66.56}{0.81} & \meanstd{70.79}{0.15} & \meanstd{34.86}{0.34} & \meanstd{51.02}{0.14} & \meanstd{63.27}{0.57} & \meanstd{74.64}{0.60} & \meanstd{47.29}{0.12} & \meanstd{60.26}{0.15} \\
\textit{SimCLR}            & \meanstd{55.96}{0.10} & \meanstd{66.87}{0.16} & \meanstd{27.43}{0.24} & \meanstd{35.10}{0.18} & \meanstd{68.85}{0.77} & \meanstd{70.73}{1.41} & \meanstd{39.04}{0.68} & \meanstd{49.12}{0.33} & \meanstd{65.53}{0.56} & \meanstd{69.26}{0.72} & \meanstd{51.36}{0.20} & \meanstd{58.22}{0.39} \\
\textit{Masked AE}         & \meanstd{48.30}{0.20} & \meanstd{61.94}{0.25} & \meanstd{20.32}{1.19} & \meanstd{32.19}{0.14} & \meanstd{63.04}{0.72} & \meanstd{69.15}{0.82} & \meanstd{39.49}{0.08} & \meanstd{48.25}{0.09} & \meanstd{66.65}{0.94} & \meanstd{66.51}{0.94} & \meanstd{47.56}{0.38} & \meanstd{55.61}{0.32} \\
\textit{Denoising AE}      & \meanstd{50.17}{0.29} & \meanstd{63.67}{0.25} & \meanstd{23.53}{0.47} & \meanstd{32.44}{0.08} & \meanstd{62.68}{0.54} & \meanstd{65.87}{0.75} & \meanstd{39.60}{0.12} & \meanstd{48.23}{0.28} & \meanstd{65.38}{0.24} & \meanstd{66.33}{0.82} & \meanstd{48.27}{0.23} & \meanstd{55.31}{0.29} \\
\textit{REBAR}             & \meanstd{36.76}{0.52} & \meanstd{59.77}{0.12} & \meanstd{16.80}{0.69} & \meanstd{30.89}{0.11} & \meanstd{53.47}{1.35} & \meanstd{65.18}{0.54} & \meanstd{24.50}{0.82} & \meanstd{46.08}{0.15} & \meanstd{53.91}{0.98} & \meanstd{68.36}{0.04} & \meanstd{37.09}{0.13} & \meanstd{54.05}{0.13} \\
\textit{RelCon}            & \meanstd{41.62}{0.10} & \meanstd{59.06}{0.33} & \meanstd{14.87}{1.34} & \meanstd{30.41}{0.14} & \meanstd{52.97}{1.03} & \meanstd{65.23}{0.33} & \meanstd{33.38}{0.47} & \meanstd{47.33}{0.03} & \meanstd{59.77}{2.83} & \meanstd{65.48}{1.57} & \meanstd{40.52}{0.98} & \meanstd{53.50}{0.40} \\
\addlinespace[1pt]
\rowcolor{gray!15}
\textbf{MorphCL}                        & \meanstd{51.68}{0.06} & \meanstd{65.34}{0.23} & \meanstd{25.88}{0.96} & \meanstd{35.16}{0.20} & \meanstd{67.41}{0.29} & \meanstd{70.02}{0.69} & \meanstd{36.52}{1.52} & \meanstd{49.55}{0.24} & \meanstd{63.03}{0.62} & \meanstd{69.40}{0.47} & \meanstd{48.90}{0.66} & \meanstd{57.90}{0.23} \\
\bottomrule
\end{tabular}%
}
\end{table*}

\begin{table*}[!ht]
\centering
\tiny
\caption{Absolute percentage point improvements on downstream linear probing results of instance-based self-supervised learning methods if combined with an additional morphological loss term. Results are reported for five downstream benchmark datasets. Pretraining was performed on both the WISDM and CAPTURE-24 dataset using different encoder backbones. Macro F1 is reported in \% as mean $\pm$ standard deviation across seeded runs.}
\label{tab:probingresults_morphimprovement}
\setlength{\tabcolsep}{3pt}
\renewcommand{\arraystretch}{1.2}
\resizebox{\textwidth}{!}{%
\begin{tabular}{@{}l*{6}{c}@{}}
\toprule
\multirow{1}{*}{}
& \multicolumn{1}{c}{WEAR}
& \multicolumn{1}{c}{Wetlab}
& \multicolumn{1}{c}{PAMAP2}
& \multicolumn{1}{c}{Hangtime}
& \multicolumn{1}{c}{RWHAR} 
& \multicolumn{1}{c}{\textit{Avg.}}\\
\midrule
\multicolumn{7}{l}{\textbf{Encoder:} DeepConvLSTM; \textbf{Pretraining:} WISDM} \\
\textit{MTL}                       & \meanstd{37.72}{0.34} & \meanstd{17.63}{0.07} & \meanstd{54.84}{0.18} & \meanstd{27.03}{0.56} & \meanstd{54.31}{0.44} & \meanstd{38.31}{0.06} \\
\textit{MTL + MorphCL}               & \meanstd{50.58}{0.34} (\textcolor{ForestGreen}{$+12.86$}) & \meanstd{23.72}{0.33} (\textcolor{ForestGreen}{$+6.09$}) & \meanstd{66.99}{0.54} (\textcolor{ForestGreen}{$+12.16$}) & \meanstd{36.03}{0.59} (\textcolor{ForestGreen}{$+9.00$}) & \meanstd{65.61}{1.12} (\textcolor{ForestGreen}{$+11.30$}) & \meanstd{48.59}{0.25} (\textcolor{ForestGreen}{$+10.28$}) \\
\textit{SimCLR}                    & \meanstd{46.74}{0.11} & \meanstd{21.21}{0.41} & \meanstd{61.47}{0.22} & \meanstd{38.64}{0.36} & \meanstd{63.29}{0.45} & \meanstd{46.27}{0.09} \\
\textit{SimCLR + MorphCL}            & \meanstd{48.41}{0.49} (\textcolor{ForestGreen}{$+1.67$}) & \meanstd{23.43}{0.42} (\textcolor{ForestGreen}{$+2.22$}) & \meanstd{63.57}{0.61} (\textcolor{ForestGreen}{$+2.10$}) & \meanstd{39.52}{0.43} (\textcolor{ForestGreen}{$+0.88$}) & \meanstd{64.33}{0.31} (\textcolor{ForestGreen}{$+1.04$}) & \meanstd{47.85}{0.26} (\textcolor{ForestGreen}{$+1.58$}) \\
\textit{REBAR}                       & \meanstd{33.64}{0.26} & \meanstd{18.16}{0.05} & \meanstd{53.86}{0.28} & \meanstd{24.25}{0.22} & \meanstd{59.15}{0.77} & \meanstd{37.81}{0.19} \\
\textit{REBAR + MorphCL}               & \meanstd{45.01}{0.25} (\textcolor{ForestGreen}{$+11.37$}) & \meanstd{25.15}{0.15} (\textcolor{ForestGreen}{$+6.99$}) & \meanstd{65.29}{0.14} (\textcolor{ForestGreen}{$+11.43$}) & \meanstd{32.62}{0.31} (\textcolor{ForestGreen}{$+8.37$}) & \meanstd{64.08}{0.71} (\textcolor{ForestGreen}{$+4.93$}) & \meanstd{46.43}{0.17} (\textcolor{ForestGreen}{$+8.62$}) \\
\textit{RelCon}                       & \meanstd{30.80}{0.11} & \meanstd{16.19}{0.11} & \meanstd{48.90}{0.63} & \meanstd{25.39}{0.24} & \meanstd{54.66}{1.23} & \meanstd{35.19}{0.36} \\
\textit{RelCon + MorphCL}               & \meanstd{44.50}{0.22} (\textcolor{ForestGreen}{$+13.70$}) & \meanstd{23.16}{0.34} (\textcolor{ForestGreen}{$+6.96$}) & \meanstd{65.25}{0.61} (\textcolor{ForestGreen}{$+16.35$}) & \meanstd{34.27}{0.26} (\textcolor{ForestGreen}{$+8.88$}) & \meanstd{62.23}{0.34} (\textcolor{ForestGreen}{$+7.57$}) & \meanstd{45.88}{0.20} (\textcolor{ForestGreen}{$+10.69$}) \\
\midrule
\multicolumn{7}{l}{\textbf{Encoder:} ViT-S; \textbf{Pretraining:} WISDM} \\
\textit{MTL}                       & \meanstd{50.05}{0.59} & \meanstd{22.39}{0.24} & \meanstd{63.84}{0.42} & \meanstd{33.71}{0.98} & \meanstd{67.03}{0.80} & \meanstd{47.41}{0.22} \\
\textit{MTL + MorphCL}               & \meanstd{53.55}{0.23} (\textcolor{ForestGreen}{$+3.50$}) & \meanstd{26.53}{0.30} (\textcolor{ForestGreen}{$+4.14$}) & \meanstd{67.17}{1.20} (\textcolor{ForestGreen}{$+3.33$}) & \meanstd{35.82}{0.51} (\textcolor{ForestGreen}{$+2.11$}) & \meanstd{69.60}{1.20} (\textcolor{ForestGreen}{$+2.57$}) & \meanstd{50.53}{0.45} (\textcolor{ForestGreen}{$+3.13$}) \\
\textit{SimCLR}                       & \meanstd{53.08}{0.62} & \meanstd{25.94}{0.41} & \meanstd{64.00}{1.25} & \meanstd{35.73}{0.44} & \meanstd{65.14}{1.44} & \meanstd{48.78}{0.57} \\
\textit{SimCLR + MorphCL}               & \meanstd{56.30}{0.21} (\textcolor{ForestGreen}{$+3.22$}) & \meanstd{25.98}{0.54} (\textcolor{ForestGreen}{$+0.04$}) & \meanstd{66.00}{0.20} (\textcolor{ForestGreen}{$+2.00$}) & \meanstd{37.36}{1.08} (\textcolor{ForestGreen}{$+1.63$}) & \meanstd{68.46}{2.54} (\textcolor{ForestGreen}{$+3.31$}) & \meanstd{50.82}{0.50} (\textcolor{ForestGreen}{$+2.04$}) \\
\textit{Masked AE}                       & \meanstd{46.58}{0.80} & \meanstd{17.43}{1.04} & \meanstd{57.86}{0.65} & \meanstd{37.19}{1.60} & \meanstd{62.32}{1.03} & \meanstd{44.28}{0.87} \\
\textit{Masked AE + MorphCL}               & \meanstd{53.20}{0.48} (\textcolor{ForestGreen}{$+6.62$}) & \meanstd{27.47}{1.01} (\textcolor{ForestGreen}{$+10.04$}) & \meanstd{66.88}{0.83} (\textcolor{ForestGreen}{$+9.02$}) & \meanstd{36.91}{0.33} (\textcolor{OrangeRed}{$-0.28$}) & \meanstd{68.37}{2.13} (\textcolor{ForestGreen}{$+6.05$}) & \meanstd{50.57}{0.50} (\textcolor{ForestGreen}{$+6.29$}) \\
\textit{Denoising AE}                       & \meanstd{49.70}{0.59} & \meanstd{20.92}{0.93} & \meanstd{62.53}{0.25} & \meanstd{38.35}{0.38} & \meanstd{67.15}{1.08} & \meanstd{47.73}{0.49} \\
\textit{Denoising AE + MorphCL}               & \meanstd{53.88}{0.11} (\textcolor{ForestGreen}{$+4.18$}) & \meanstd{25.38}{0.27} (\textcolor{ForestGreen}{$+4.45$}) & \meanstd{63.80}{0.31} (\textcolor{ForestGreen}{$+1.27$}) & \meanstd{37.23}{0.45} (\textcolor{OrangeRed}{$-1.12$}) & \meanstd{66.30}{1.09} (\textcolor{OrangeRed}{$-0.85$}) & \meanstd{49.32}{0.33} (\textcolor{ForestGreen}{$+1.59$}) \\
\textit{REBAR}                       & \meanstd{42.59}{0.68} & \meanstd{16.80}{0.40} & \meanstd{59.01}{0.38} & \meanstd{29.12}{0.38} & \meanstd{59.89}{2.57} & \meanstd{41.48}{0.68} \\
\textit{REBAR + MorphCL}               & \meanstd{52.77}{0.47} (\textcolor{ForestGreen}{$+10.18$}) & \meanstd{26.20}{1.25} (\textcolor{ForestGreen}{$+9.40$}) & \meanstd{65.13}{0.45} (\textcolor{ForestGreen}{$+6.11$}) & \meanstd{36.45}{1.07} (\textcolor{ForestGreen}{$+7.33$}) & \meanstd{68.48}{0.88} (\textcolor{ForestGreen}{$+8.59$}) & \meanstd{49.81}{0.39} (\textcolor{ForestGreen}{$+8.32$}) \\
\textit{RelCon}                       & \meanstd{37.53}{0.63} & \meanstd{13.70}{0.61} & \meanstd{43.71}{1.19} & \meanstd{25.88}{0.26} & \meanstd{52.21}{1.18} & \meanstd{34.61}{0.33} \\
\textit{RelCon + MorphCL}               & \meanstd{50.91}{0.50} (\textcolor{ForestGreen}{$+13.37$}) & \meanstd{24.93}{0.46} (\textcolor{ForestGreen}{$+11.23$}) & \meanstd{66.04}{0.74} (\textcolor{ForestGreen}{$+22.34$}) & \meanstd{39.30}{0.21} (\textcolor{ForestGreen}{$+13.42$}) & \meanstd{68.62}{2.13} (\textcolor{ForestGreen}{$+16.41$}) & \meanstd{49.96}{0.40} (\textcolor{ForestGreen}{$+15.36$}) \\
\midrule
\multicolumn{7}{l}{\textbf{Encoder:} ViT-S; \textbf{Pretraining:} CAPTURE-24} \\
\textit{MTL}                       & \meanstd{49.34}{0.74} & \meanstd{22.44}{0.20} & \meanstd{66.56}{0.81} & \meanstd{34.86}{0.34} & \meanstd{63.27}{0.57} & \meanstd{47.29}{0.12} \\
\textit{MTL + MorphCL}               & \meanstd{56.53}{0.29} (\textcolor{ForestGreen}{$+7.19$}) & \meanstd{26.65}{0.74} (\textcolor{ForestGreen}{$+4.22$}) & \meanstd{69.70}{0.65} (\textcolor{ForestGreen}{$+3.14$}) & \meanstd{38.90}{1.01} (\textcolor{ForestGreen}{$+4.03$}) & \meanstd{68.83}{0.86} (\textcolor{ForestGreen}{$+5.56$}) & \meanstd{52.12}{0.33} (\textcolor{ForestGreen}{$+4.83$}) \\
\textit{SimCLR}                       & \meanstd{55.96}{0.10} & \meanstd{27.43}{0.24} & \meanstd{68.85}{0.77} & \meanstd{39.04}{0.68} & \meanstd{65.53}{0.56} & \meanstd{51.36}{0.20} \\
\textit{SimCLR + MorphCL}               & \meanstd{56.67}{0.24} (\textcolor{ForestGreen}{$+0.71$}) & \meanstd{26.93}{0.24} (\textcolor{OrangeRed}{$-0.50$}) & \meanstd{68.29}{0.29} (\textcolor{OrangeRed}{$-0.56$}) & \meanstd{39.02}{1.53} (\textcolor{OrangeRed}{$-0.02$}) & \meanstd{63.94}{1.07} (\textcolor{OrangeRed}{$-1.59$}) & \meanstd{50.97}{0.23} (\textcolor{OrangeRed}{$-0.39$}) \\
\textit{Masked AE}                       & \meanstd{48.30}{0.20} & \meanstd{20.32}{1.19} & \meanstd{63.04}{0.72} & \meanstd{39.49}{0.08} & \meanstd{66.65}{0.94} & \meanstd{47.56}{0.38} \\
\textit{Masked AE + MorphCL}               & \meanstd{54.21}{0.03} (\textcolor{ForestGreen}{$+5.91$}) & \meanstd{27.08}{1.07} (\textcolor{ForestGreen}{$+6.76$}) & \meanstd{65.55}{0.74} (\textcolor{ForestGreen}{$+2.52$}) & \meanstd{37.13}{0.95} (\textcolor{OrangeRed}{$-2.36$}) & \meanstd{66.26}{0.27} (\textcolor{OrangeRed}{$-0.40$}) & \meanstd{50.05}{0.37} (\textcolor{ForestGreen}{$+2.49$}) \\
\textit{Denoising AE}                       & \meanstd{50.17}{0.29} & \meanstd{23.53}{0.47} & \meanstd{62.68}{0.54} & \meanstd{39.60}{0.12} & \meanstd{65.38}{0.24} & \meanstd{48.27}{0.23} \\
\textit{Denoising AE + MorphCL}               & \meanstd{56.41}{0.59} (\textcolor{ForestGreen}{$+6.24$}) & \meanstd{25.96}{0.64} (\textcolor{ForestGreen}{$+2.43$}) & \meanstd{67.23}{0.57} (\textcolor{ForestGreen}{$+4.55$}) & \meanstd{38.72}{1.33} (\textcolor{OrangeRed}{$-0.87$}) & \meanstd{67.32}{1.25} (\textcolor{ForestGreen}{$+1.94$}) & \meanstd{51.13}{0.33} (\textcolor{ForestGreen}{$+2.86$}) \\
\textit{REBAR}                       & \meanstd{36.76}{0.52} & \meanstd{16.80}{0.69} & \meanstd{53.47}{1.35} & \meanstd{24.50}{0.82} & \meanstd{53.91}{0.98} & \meanstd{37.09}{0.13} \\
\textit{REBAR + MorphCL}               & \meanstd{50.83}{0.48} (\textcolor{ForestGreen}{$+14.07$}) & \meanstd{25.39}{1.40} (\textcolor{ForestGreen}{$+8.59$}) & \meanstd{66.97}{0.25} (\textcolor{ForestGreen}{$+13.51$}) & \meanstd{38.50}{0.56} (\textcolor{ForestGreen}{$+14.00$}) & \meanstd{59.67}{1.74} (\textcolor{ForestGreen}{$+5.75$}) & \meanstd{48.27}{0.49} (\textcolor{ForestGreen}{$+11.18$}) \\
\textit{RelCon}                       & \meanstd{41.62}{0.10} & \meanstd{14.87}{1.34} & \meanstd{52.97}{1.03} & \meanstd{33.38}{0.47} & \meanstd{59.77}{2.83} & \meanstd{40.52}{0.98} \\
\textit{RelCon + MorphCL}               & \meanstd{54.19}{0.72} (\textcolor{ForestGreen}{$+12.58$}) & \meanstd{27.29}{0.43} (\textcolor{ForestGreen}{$+12.42$}) & \meanstd{64.35}{0.45} (\textcolor{ForestGreen}{$+11.38$}) & \meanstd{37.13}{0.86} (\textcolor{ForestGreen}{$+3.74$}) & \meanstd{66.69}{0.99} (\textcolor{ForestGreen}{$+6.92$}) & \meanstd{49.93}{0.30} (\textcolor{ForestGreen}{$+9.41$}) \\
\bottomrule
\end{tabular}%
}
\end{table*}

\begin{table*}[!ht]
\centering
\tiny
\caption{Absolute finetuning percentage point improvement of existing SSL methods if combined with MorphCL. Results are seed-averaged across 3 runs. All frameworks benefit from being combined with MorphCL with the MTL framework being the best performing method. Macro F1 is reported in \% as mean $\pm$ standard deviation across seeded runs.}
\label{tab:finetuningresults_morphimprovement}
\setlength{\tabcolsep}{3pt}
\renewcommand{\arraystretch}{1.2}
\resizebox{\textwidth}{!}{%
\begin{tabular}{@{}l*{6}{c}@{}}
\toprule
\multirow{1}{*}{}
& \multicolumn{1}{c}{WEAR}
& \multicolumn{1}{c}{Wetlab}
& \multicolumn{1}{c}{PAMAP2}
& \multicolumn{1}{c}{Hangtime}
& \multicolumn{1}{c}{RWHAR} 
& \multicolumn{1}{c}{\textit{Avg.}}\\
\midrule
\multicolumn{7}{l}{\textbf{Encoder:} DeepConvLSTM; \textbf{Pretraining:} WISDM} \\
\textit{MTL}                       & \meanstd{57.86}{0.53} & \meanstd{27.99}{0.19} & \meanstd{69.69}{0.16} & \meanstd{48.59}{0.23} & \meanstd{66.77}{1.46} & \meanstd{54.18}{0.44} \\
\textit{MTL + MorphCL}               & \meanstd{62.59}{0.46} (\textcolor{ForestGreen}{$+4.74$}) & \meanstd{33.36}{0.17} (\textcolor{ForestGreen}{$+5.37$}) & \meanstd{72.37}{1.04} (\textcolor{ForestGreen}{$+2.67$}) & \meanstd{52.99}{0.54} (\textcolor{ForestGreen}{$+4.40$}) & \meanstd{69.06}{0.92} (\textcolor{ForestGreen}{$+2.29$}) & \meanstd{58.07}{0.22} (\textcolor{ForestGreen}{$+3.89$}) \\
\textit{SimCLR}                       & \meanstd{60.66}{0.25} & \meanstd{32.58}{0.25} & \meanstd{73.64}{0.80} & \meanstd{52.40}{0.12} & \meanstd{67.35}{0.89} & \meanstd{57.32}{0.31} \\
\textit{SimCLR + MorphCL}               & \meanstd{62.33}{0.28} (\textcolor{ForestGreen}{$+1.67$}) & \meanstd{33.46}{0.30} (\textcolor{ForestGreen}{$+0.88$}) & \meanstd{71.68}{0.97} (\textcolor{OrangeRed}{$-1.96$}) & \meanstd{53.15}{0.22} (\textcolor{ForestGreen}{$+0.75$}) & \meanstd{67.21}{1.41} (\textcolor{OrangeRed}{$-0.14$}) & \meanstd{57.57}{0.28} (\textcolor{ForestGreen}{$+0.24$}) \\
\textit{REBAR}                       & \meanstd{56.64}{0.20} & \meanstd{27.76}{0.08} & \meanstd{67.06}{0.64} & \meanstd{50.89}{0.39} & \meanstd{63.81}{0.75} & \meanstd{53.23}{0.29} \\
\textit{REBAR + MorphCL}               & \meanstd{59.52}{0.62} (\textcolor{ForestGreen}{$+2.88$}) & \meanstd{31.88}{0.34} (\textcolor{ForestGreen}{$+4.12$}) & \meanstd{69.79}{0.07} (\textcolor{ForestGreen}{$+2.73$}) & \meanstd{52.62}{0.20} (\textcolor{ForestGreen}{$+1.73$}) & \meanstd{71.23}{0.37} (\textcolor{ForestGreen}{$+7.42$}) & \meanstd{57.01}{0.04} (\textcolor{ForestGreen}{$+3.78$}) \\
\textit{RelCon}                       & \meanstd{56.43}{0.15} & \meanstd{27.22}{0.54} & \meanstd{65.24}{1.01} & \meanstd{50.87}{0.23} & \meanstd{64.37}{1.24} & \meanstd{52.82}{0.04} \\
\textit{RelCon + MorphCL}               & \meanstd{60.21}{0.11} (\textcolor{ForestGreen}{$+3.78$}) & \meanstd{31.98}{0.29} (\textcolor{ForestGreen}{$+4.76$}) & \meanstd{69.50}{2.62} (\textcolor{ForestGreen}{$+4.26$}) & \meanstd{52.40}{0.21} (\textcolor{ForestGreen}{$+1.53$}) & \meanstd{66.79}{0.62} (\textcolor{ForestGreen}{$+2.42$}) & \meanstd{56.18}{0.40} (\textcolor{ForestGreen}{$+3.35$}) \\
\midrule
\multicolumn{7}{l}{\textbf{Encoder:} ViT-S; \textbf{Pretraining:} WISDM} \\
\textit{MTL}                       & \meanstd{66.02}{0.33} & \meanstd{33.25}{0.30} & \meanstd{70.05}{0.70} & \meanstd{49.67}{0.15} & \meanstd{68.58}{0.45} & \meanstd{57.51}{0.27} \\
\textit{MTL + MorphCL}               & \meanstd{66.62}{0.24} (\textcolor{ForestGreen}{$+0.60$}) & \meanstd{34.46}{0.24} (\textcolor{ForestGreen}{$+1.21$}) & \meanstd{70.82}{0.16} (\textcolor{ForestGreen}{$+0.77$}) & \meanstd{49.30}{0.09} (\textcolor{OrangeRed}{$-0.38$}) & \meanstd{69.93}{0.22} (\textcolor{ForestGreen}{$+1.35$}) & \meanstd{58.23}{0.16} (\textcolor{ForestGreen}{$+0.71$}) \\
\textit{SimCLR}                       & \meanstd{63.12}{0.41} & \meanstd{32.71}{0.24} & \meanstd{67.35}{0.87} & \meanstd{47.96}{0.30} & \meanstd{65.25}{0.85} & \meanstd{55.28}{0.25} \\
\textit{SimCLR + MorphCL}               & \meanstd{64.23}{0.05} (\textcolor{ForestGreen}{$+1.11$}) & \meanstd{33.69}{0.03} (\textcolor{ForestGreen}{$+0.98$}) & \meanstd{68.62}{0.34} (\textcolor{ForestGreen}{$+1.28$}) & \meanstd{48.78}{0.47} (\textcolor{ForestGreen}{$+0.82$}) & \meanstd{65.93}{1.15} (\textcolor{ForestGreen}{$+0.68$}) & \meanstd{56.25}{0.19} (\textcolor{ForestGreen}{$+0.97$}) \\
\textit{Masked AE}                       & \meanstd{60.57}{0.07} & \meanstd{31.09}{0.08} & \meanstd{66.59}{1.82} & \meanstd{47.65}{0.14} & \meanstd{65.92}{0.16} & \meanstd{54.36}{0.32} \\
\textit{Masked AE + MorphCL}               & \meanstd{62.98}{0.12} (\textcolor{ForestGreen}{$+2.42$}) & \meanstd{32.98}{0.28} (\textcolor{ForestGreen}{$+1.89$}) & \meanstd{70.20}{0.29} (\textcolor{ForestGreen}{$+3.61$}) & \meanstd{48.17}{0.24} (\textcolor{ForestGreen}{$+0.52$}) & \meanstd{66.55}{0.87} (\textcolor{ForestGreen}{$+0.63$}) & \meanstd{56.18}{0.16} (\textcolor{ForestGreen}{$+1.81$}) \\
\textit{Denoising AE}                       & \meanstd{60.51}{0.08} & \meanstd{31.25}{0.16} & \meanstd{65.63}{1.36} & \meanstd{47.21}{0.24} & \meanstd{64.11}{0.64} & \meanstd{53.74}{0.24} \\
\textit{Denoising AE + MorphCL}               & \meanstd{62.88}{0.26} (\textcolor{ForestGreen}{$+2.37$}) & \meanstd{32.37}{0.16} (\textcolor{ForestGreen}{$+1.12$}) & \meanstd{68.72}{0.89} (\textcolor{ForestGreen}{$+3.08$}) & \meanstd{48.11}{0.23} (\textcolor{ForestGreen}{$+0.90$}) & \meanstd{64.28}{0.99} (\textcolor{ForestGreen}{$+0.18$}) & \meanstd{55.27}{0.13} (\textcolor{ForestGreen}{$+1.53$}) \\
\textit{REBAR}                       & \meanstd{58.84}{0.06} & \meanstd{30.13}{0.30} & \meanstd{64.95}{1.14} & \meanstd{46.70}{0.33} & \meanstd{64.11}{0.68} & \meanstd{52.95}{0.21} \\
\textit{REBAR + MorphCL}               & \meanstd{61.95}{0.07} (\textcolor{ForestGreen}{$+3.11$}) & \meanstd{32.37}{0.12} (\textcolor{ForestGreen}{$+2.24$}) & \meanstd{67.01}{0.62} (\textcolor{ForestGreen}{$+2.06$}) & \meanstd{48.46}{0.16} (\textcolor{ForestGreen}{$+1.76$}) & \meanstd{64.33}{0.78} (\textcolor{ForestGreen}{$+0.22$}) & \meanstd{54.83}{0.24} (\textcolor{ForestGreen}{$+1.88$}) \\
\textit{RelCon}                       & \meanstd{59.42}{0.16} & \meanstd{30.20}{0.11} & \meanstd{64.54}{0.95} & \meanstd{47.30}{0.01} & \meanstd{65.69}{0.87} & \meanstd{53.43}{0.30} \\
\textit{RelCon + MorphCL}               & \meanstd{60.78}{0.05} (\textcolor{ForestGreen}{$+1.37$}) & \meanstd{31.50}{0.25} (\textcolor{ForestGreen}{$+1.31$}) & \meanstd{65.43}{0.55} (\textcolor{ForestGreen}{$+0.89$}) & \meanstd{47.42}{0.12} (\textcolor{ForestGreen}{$+0.11$}) & \meanstd{64.64}{0.54} (\textcolor{OrangeRed}{$-1.05$}) & \meanstd{53.95}{0.15} (\textcolor{ForestGreen}{$+0.52$}) \\
\midrule
\multicolumn{7}{l}{\textbf{Encoder:} ViT-S; \textbf{Pretraining:} CAPTURE-24} \\
\textit{MTL}                       & \meanstd{69.07}{0.12} & \meanstd{35.79}{0.04} & \meanstd{70.79}{0.15} & \meanstd{51.02}{0.14} & \meanstd{74.64}{0.60} & \meanstd{60.26}{0.15} \\
\textit{MTL + MorphCL}               & \meanstd{70.47}{0.30} (\textcolor{ForestGreen}{$+1.39$}) & \meanstd{36.26}{0.04} (\textcolor{ForestGreen}{$+0.47$}) & \meanstd{72.04}{0.25} (\textcolor{ForestGreen}{$+1.25$}) & \meanstd{51.58}{0.25} (\textcolor{ForestGreen}{$+0.56$}) & \meanstd{73.07}{0.49} (\textcolor{OrangeRed}{$-1.57$}) & \meanstd{60.68}{0.11} (\textcolor{ForestGreen}{$+0.42$}) \\
\textit{SimCLR}                       & \meanstd{66.87}{0.16} & \meanstd{35.10}{0.18} & \meanstd{70.73}{1.41} & \meanstd{49.12}{0.33} & \meanstd{69.26}{0.72} & \meanstd{58.22}{0.39} \\
\textit{SimCLR + MorphCL}               & \meanstd{67.32}{0.27} (\textcolor{ForestGreen}{$+0.45$}) & \meanstd{35.76}{0.26} (\textcolor{ForestGreen}{$+0.67$}) & \meanstd{72.07}{0.20} (\textcolor{ForestGreen}{$+1.34$}) & \meanstd{50.26}{0.39} (\textcolor{ForestGreen}{$+1.14$}) & \meanstd{68.20}{0.55} (\textcolor{OrangeRed}{$-1.06$}) & \meanstd{58.72}{0.19} (\textcolor{ForestGreen}{$+0.51$}) \\
\textit{Masked AE}                       & \meanstd{61.94}{0.25} & \meanstd{32.19}{0.14} & \meanstd{69.15}{0.82} & \meanstd{48.25}{0.09} & \meanstd{66.51}{0.94} & \meanstd{55.61}{0.32} \\
\textit{Masked AE + MorphCL}               & \meanstd{66.55}{0.25} (\textcolor{ForestGreen}{$+4.60$}) & \meanstd{35.06}{0.21} (\textcolor{ForestGreen}{$+2.87$}) & \meanstd{69.78}{0.72} (\textcolor{ForestGreen}{$+0.62$}) & \meanstd{49.66}{0.27} (\textcolor{ForestGreen}{$+1.41$}) & \meanstd{68.87}{0.33} (\textcolor{ForestGreen}{$+2.36$}) & \meanstd{57.98}{0.13} (\textcolor{ForestGreen}{$+2.37$}) \\
\textit{Denoising AE}                       & \meanstd{63.67}{0.25} & \meanstd{32.44}{0.08} & \meanstd{65.87}{0.75} & \meanstd{48.23}{0.28} & \meanstd{66.33}{0.82} & \meanstd{55.31}{0.29} \\
\textit{Denoising AE + MorphCL}               & \meanstd{68.00}{0.33} (\textcolor{ForestGreen}{$+4.33$}) & \meanstd{35.11}{0.16} (\textcolor{ForestGreen}{$+2.67$}) & \meanstd{69.66}{0.64} (\textcolor{ForestGreen}{$+3.79$}) & \meanstd{50.67}{0.22} (\textcolor{ForestGreen}{$+2.44$}) & \meanstd{68.30}{0.61} (\textcolor{ForestGreen}{$+1.97$}) & \meanstd{58.35}{0.25} (\textcolor{ForestGreen}{$+3.04$}) \\
\textit{REBAR}                       & \meanstd{59.77}{0.12} & \meanstd{30.89}{0.11} & \meanstd{65.18}{0.54} & \meanstd{46.08}{0.15} & \meanstd{68.36}{0.04} & \meanstd{54.05}{0.13} \\
\textit{REBAR + MorphCL}               & \meanstd{66.13}{0.41} (\textcolor{ForestGreen}{$+6.35$}) & \meanstd{34.93}{0.36} (\textcolor{ForestGreen}{$+4.04$}) & \meanstd{70.79}{0.34} (\textcolor{ForestGreen}{$+5.62$}) & \meanstd{49.99}{0.45} (\textcolor{ForestGreen}{$+3.91$}) & \meanstd{68.76}{0.27} (\textcolor{ForestGreen}{$+0.40$}) & \meanstd{58.12}{0.16} (\textcolor{ForestGreen}{$+4.07$}) \\
\textit{RelCon}                       & \meanstd{59.06}{0.33} & \meanstd{30.41}{0.14} & \meanstd{65.23}{0.33} & \meanstd{47.33}{0.03} & \meanstd{65.48}{1.57} & \meanstd{53.50}{0.40} \\
\textit{RelCon + MorphCL}               & \meanstd{64.43}{0.36} (\textcolor{ForestGreen}{$+5.37$}) & \meanstd{33.82}{0.21} (\textcolor{ForestGreen}{$+3.42$}) & \meanstd{69.42}{1.80} (\textcolor{ForestGreen}{$+4.19$}) & \meanstd{48.57}{0.26} (\textcolor{ForestGreen}{$+1.25$}) & \meanstd{67.05}{0.16} (\textcolor{ForestGreen}{$+1.57$}) & \meanstd{56.66}{0.42} (\textcolor{ForestGreen}{$+3.16$}) \\
\bottomrule
\end{tabular}%
}
\end{table*}

\subsection{Sensitivity Analysis on Motif Discovery Parameters}

We investigate the sensitivity of our motif discovery on the WISDM dataset to hyperparameters along two dimensions. First, Table~\ref{tab:hyperparameter_sweep_sigma} reports the discovered motifs for different motif distance thresholds $\sigma$. Second, in Table~\ref{tab:hyperparameter_sweep_k} we ablate different numbers of minimum occurrences $r$ of a motif to be considered a motif. Note that we do not specifically ablate on motif sizes larger than 0.2s as we argue that (1) larger motifs can be considered a subset of these and (2) smaller motifs are too small given our fixed sampling rate of 50 Hz (less than 10 samples) to contain any meaningful motion. One can see that motif discovery remains relatively stable across differently chosen $\sigma$ and $r$ values. For smaller $\sigma$, we see less motifs and windows extracted, as fewer motif candidates are considered close enough to each other. For larger distances, number of extracted windows slightly decreases, which we attribute to our applied masking strategy of additional motif candidates ($>k$) as detailed in the paper. As we apply motif discovery in one hour chunks to both CAPTURE-24 and WISDM, we expect results below computed on WISDM to be representative for CAPTURE-24 as well. Downstream results (MTL + MorphCL) show our MorphCL framework being stable across different motif discovery hyperparameters. Note that we apply our initial configuration for clustering ($\rho=100$ and $\text{support}=100$) during these experiments. Our chosen configuration of $\sigma=1.0$ and $r=100$ returns most stable results for motifs of length 0.2 seconds.

\begin{table}[!ht]
    \centering
    \tiny
    \caption{Hyperparameter sweep for $\sigma$ with fixed $r=100$ for a motif length of 0.2s.}
    \label{tab:hyperparameter_sweep_sigma}
    \begin{tabular}{@{}rrrrr@{}}
        \toprule
        $\sigma$ & Discovered motifs & Extracted windows & Relative to full dataset & Downstream F1 \\
        \midrule
        $0.5$ & $636$ & $15,816$ & $31.69\%$ & \meanstd{49.61}{0.64} \\
        $1.0$ & $758$ & $21,318$ & $42.71\%$ & \meanstd{50.53}{0.45} \\
        $1.5$ & $788$ & $23,202$ & $46.49\%$ & \meanstd{49.89}{0.61} \\
        $2.0$ & $690$ & $20,793$ & $41.66\%$ & \meanstd{49.67}{0.58} \\
        \bottomrule
    \end{tabular}
\end{table}

\begin{table}[!ht]
    \centering
    \tiny
    \caption{Hyperparameter sweep for $r$ with fixed $\sigma=1.0$ for a motif length of 0.2s.}
    \label{tab:hyperparameter_sweep_k}
    \begin{tabular}{@{}rrrrr@{}}
        \toprule
        $r$ & Discovered motifs & Extracted windows & Relative to full dataset & Downstream F1 \\
        \midrule
         $50$ & $1,145$ & $19,943$ & $40.00\%$ & \meanstd{49.95}{0.75} \\
        $100$ &   $758$ & $21,318$ & $42.71\%$ & \meanstd{50.53}{0.45} \\
        $150$ &   $656$ & $22,899$ & $45.88\%$ & \meanstd{49.88}{0.75} \\
        $200$ &   $544$ & $22,956$ & $45.99\%$ & \meanstd{49.81}{0.55} \\
        \bottomrule
    \end{tabular}
\end{table}

\subsection{Sensitivity Analysis on Clustering Parameters}

We ablate $\rho$ and cluster support on the WISDM dataset. To investigate the stability across changes in hyperparameters we report DBCV (Density-Based Clustering Validation), a cluster-validity measure as suggested by~\citet{moulaviDensityBased2014} that does not require ground-truth labels ($[-1,1]$, 1 being well-separated clusters) and AMI (Adjusted Mutual Information) against activity labels  ($[0,1]$ 1 being perfect agreement between clusters and activity labels) alongside cluster count and noise rate. Both DBCV and AMI remain stable across $\rho \in [50, 200]$ and support $\in [50, 200]$, indicating the method is not narrowly tuned to a single operating point. At tested extremes ($400$), quality predictably degrades as excessively high $\rho$ exceeds the WISDM dataset’s density scale, while excessively high cluster support forces merging of distinct sub-clusters into a few coarse groups. Downstream results (MTL + MorphCL) show that our MorphCL framework is stable across different clustering hyperparameters. Our chosen configuration of $\rho=100$ and support$=100$ returns the most stable results.

\begin{table}[!ht]
    \centering
    \tiny
    \caption{Hyperparameter sweep for $\rho$ with fixed support $= 100$.}
    \label{tab:changing_rho}
    \begin{tabular}{@{}rrrrrr@{}}
        \toprule
        $\rho$ & $n_{\text{clusters}}$ & Noise & DBCV & AMI (vs. activity) & Downstream F1 \\
        \midrule
         $50$ & $19$ & $74.70\%$ & $0.3220$ & $0.6954$ & \meanstd{50.74}{0.60} \\
        $100$ &  $8$ & $78.83\%$ & $0.2434$ & $0.7122$ & \meanstd{50.53}{0.45} \\
        $200$ &  $6$ & $82.38\%$ & $0.3287$ & $0.6595$ & \meanstd{50.01}{0.74} \\
        $400$ & --   & $100\%$   & --       & --       & --                \\
        \bottomrule
    \end{tabular}
\end{table}

\begin{table}[!ht]
    \centering
    \tiny
    \caption{Hyperparameter sweep support with fixed $\rho=100$.}
    \label{tab:changing_support}
    \begin{tabular}{@{}rrrrrr@{}}
        \toprule
        Support & $n_{\text{clusters}}$ & Noise & DBCV & AMI (vs. activity) & Downstream F1 \\
        \midrule
         $50$ & $8$ & $78.83\%$ & $0.2434$ & $0.7122$ & \meanstd{50.73}{0.60} \\
        $100$ & $8$ & $78.83\%$ & $0.2434$ & $0.7122$ & \meanstd{50.53}{0.45} \\
        $200$ & $6$ & $80.16\%$ & $0.2245$ & $0.6674$ & \meanstd{49.67}{0.80} \\
        $400$ & $2$ & $14.64\%$ & $0.0622$ & $0.0357$ & \meanstd{49.28}{0.47} \\
        \bottomrule
    \end{tabular}
\end{table}

\subsection{Subset Evaluation}

To separate the contribution of the MorphCL loss from that of the data filtering, we train the base objectives (MTL, SimCLR, Denoising AE, Masked AE, REBAR and RelCon) directly on the motif-filtered subset ("Motif") and the morphological-clustered subset using the same balanced sampler as MorphCL ("Morph"), with and without the MorphCL loss. We also test random subsets matched in size and repeated samples, to isolate the effect of size alone. Numbers are averaged over the five downstream tasks across three seeds. 

On WISDM (see Table~\ref{tab:subset_wisdm}), filtering alone results in comparable results to using the full dataset, with few exceptions slightly improving or degrading in performance. Adding MorphCL improves results across all subsets, demonstrating that the gain is attributable to the objective, not the subset. On CAPTURE-24 (see Table~\ref{tab:subset_cap24}), motif-based filtering has a positive effect on SimCLR, MTL and REBAR and a negative effect on RelCon. Nevertheless, MorphCL adds a comparable further gain to all frameworks. For example, for MTL, motif filtering contributes a gain of $+2.44$ and adding MorphCL on the same subset achieves $+2.89$ more, for a total improvement of $+5.33$ over the full-corpus baseline. Filtering and the MorphCL loss each contribute roughly half the total gain, confirming that gains reported in the main paper are not merely on the data selection.

\begin{table}[!ht]
    \centering
    \tiny
    \setlength{\tabcolsep}{3pt}
    \renewcommand{\arraystretch}{1.2}
    \caption{Downstream F1 scores (\%; mean $\pm$ standard deviation) for different subset pretraining on the WISDM dataset.}
    \label{tab:subset_wisdm}
    \begin{tabular}{@{}lccccc@{}}
        \toprule
        Method & Full & Motif & Morph & Random ($N$=Motif) & Random ($N$=Morph) \\
        \midrule
        SimCLR
        & \meanstd{48.78}{0.57}
        & \meanstd{47.75}{0.62}
        & \meanstd{47.88}{0.62}
        & \meanstd{46.69}{0.56}
        & \meanstd{46.29}{0.43} \\
        \textbf{SimCLR + MorphCL}
        & \meanstd{\mathbf{50.82}}{0.50}
        & \meanstd{\mathbf{50.53}}{0.69}
        & \meanstd{\mathbf{49.80}}{0.50}
        & \meanstd{\mathbf{49.51}}{0.36}
        & \meanstd{\mathbf{49.16}}{0.58} \\

        MTL
        & \meanstd{47.41}{0.22}
        & \meanstd{46.13}{0.36}
        & \meanstd{46.86}{0.36}
        & \meanstd{45.80}{0.42}
        & \meanstd{45.96}{0.41} \\
        \textbf{MTL + MorphCL}
        & \meanstd{\mathbf{50.53}}{0.45}
        & \meanstd{\mathbf{49.70}}{0.79}
        & \meanstd{\mathbf{49.42}}{0.49}
        & \meanstd{\mathbf{49.46}}{0.59}
        & \meanstd{\mathbf{48.50}}{0.41} \\

        Denoising AE
        & \meanstd{47.73}{0.49}
        & \meanstd{47.11}{0.71}
        & \meanstd{48.68}{0.48}
        & \meanstd{46.64}{0.66}
        & \meanstd{45.94}{0.64} \\
        \textbf{Denoising AE + MorphCL}
        & \meanstd{\mathbf{49.32}}{0.33}
        & \meanstd{\mathbf{49.75}}{0.59}
        & \meanstd{\mathbf{50.25}}{0.74}
        & \meanstd{\mathbf{49.47}}{0.86}
        & \meanstd{\mathbf{49.68}}{0.71} \\

        Masked AE
        & \meanstd{44.28}{0.87}
        & \meanstd{43.56}{0.66}
        & \meanstd{42.42}{0.87}
        & \meanstd{43.54}{0.64}
        & \meanstd{42.76}{0.53} \\
        \textbf{Masked AE + MorphCL}
        & \meanstd{\mathbf{50.57}}{0.50}
        & \meanstd{\mathbf{50.06}}{0.47}
        & \meanstd{\mathbf{50.12}}{0.88}
        & \meanstd{\mathbf{50.07}}{0.65}
        & \meanstd{\mathbf{50.08}}{0.64} \\

        REBAR
        & \meanstd{41.48}{0.68}
        & \meanstd{43.06}{0.90}
        & \meanstd{41.63}{0.41}
        & \meanstd{39.85}{0.98}
        & \meanstd{42.17}{0.79} \\
        \textbf{REBAR + MorphCL}
        & \meanstd{\mathbf{49.81}}{0.39}
        & \meanstd{\mathbf{50.17}}{1.01}
        & \meanstd{\mathbf{50.22}}{0.54}
        & \meanstd{\mathbf{50.22}}{0.54}
        & \meanstd{\mathbf{50.22}}{0.54} \\

        RelCon
        & \meanstd{34.61}{0.33}
        & \meanstd{34.41}{0.71}
        & \meanstd{36.40}{1.18}
        & \meanstd{36.72}{0.83}
        & \meanstd{32.20}{1.19} \\
        \textbf{RelCon + MorphCL}
        & \meanstd{\mathbf{49.96}}{0.40}
        & \meanstd{\mathbf{48.83}}{0.56}
        & \meanstd{\mathbf{48.83}}{0.56}
        & \meanstd{\mathbf{48.83}}{0.56}
        & \meanstd{\mathbf{48.83}}{0.56} \\
        \bottomrule
    \end{tabular}
\end{table}

\begin{table}[!ht]
    \centering
    \tiny
    \setlength{\tabcolsep}{3pt}
    \renewcommand{\arraystretch}{1.2}
    \caption{Downstream F1 scores (\%; mean $\pm$ standard deviation) for different subset pretraining on the CAPTURE-24 dataset.}
    \label{tab:subset_cap24}
    \begin{tabular}{@{}lccccc@{}}
        \toprule
        Method & Full & Motif & Morph & Random ($N$=Motif) & Random ($N$=Morph) \\
        \midrule
        SimCLR
        & \meanstd{\mathbf{51.36}}{0.20}
        & \meanstd{52.05}{0.47}
        & \meanstd{45.93}{0.44}
        & \meanstd{50.14}{0.47}
        & \meanstd{46.51}{0.46} \\
        \textbf{SimCLR + MorphCL}
        & \meanstd{50.97}{0.23}
        & \meanstd{\mathbf{52.16}}{0.48}
        & \meanstd{\mathbf{48.12}}{0.80}
        & \meanstd{\mathbf{51.34}}{0.45}
        & \meanstd{\mathbf{49.26}}{0.43} \\

        MTL
        & \meanstd{47.29}{0.12}
        & \meanstd{49.73}{0.85}
        & \meanstd{49.48}{0.59}
        & \meanstd{45.08}{0.47}
        & \meanstd{46.61}{0.53} \\
        \textbf{MTL + MorphCL}
        & \meanstd{\mathbf{52.12}}{0.33}
        & \meanstd{\mathbf{52.62}}{0.52}
        & \meanstd{\mathbf{51.41}}{0.46}
        & \meanstd{\mathbf{50.92}}{0.41}
        & \meanstd{\mathbf{50.61}}{0.60} \\

        Denoising AE
        & \meanstd{48.27}{0.23}
        & \meanstd{46.65}{0.42}
        & \meanstd{46.21}{0.58}
        & \meanstd{46.62}{0.57}
        & \meanstd{44.77}{0.68} \\
        \textbf{Denoising AE + MorphCL}
        & \meanstd{\mathbf{51.13}}{0.33}
        & \meanstd{\mathbf{50.26}}{0.61}
        & \meanstd{\mathbf{49.11}}{0.51}
        & \meanstd{\mathbf{49.90}}{0.67}
        & \meanstd{\mathbf{49.57}}{0.41} \\

        Masked AE
        & \meanstd{47.56}{0.38}
        & \meanstd{47.39}{0.65}
        & \meanstd{46.46}{0.49}
        & \meanstd{46.20}{0.68}
        & \meanstd{45.60}{0.77} \\
        \textbf{Masked AE + MorphCL}
        & \meanstd{\mathbf{50.05}}{0.37}
        & \meanstd{\mathbf{49.28}}{0.42}
        & \meanstd{\mathbf{48.83}}{0.65}
        & \meanstd{\mathbf{49.57}}{0.38}
        & \meanstd{\mathbf{49.22}}{0.41} \\

        REBAR
        & \meanstd{37.09}{0.13}
        & \meanstd{42.63}{0.62}
        & \meanstd{42.53}{0.87}
        & \meanstd{36.39}{0.67}
        & \meanstd{32.21}{0.61} \\
        \textbf{REBAR + MorphCL}
        & \meanstd{\mathbf{48.27}}{0.49}
        & \meanstd{\mathbf{48.86}}{0.63}
        & \meanstd{\mathbf{49.22}}{0.41}
        & \meanstd{\mathbf{48.03}}{0.43}
        & \meanstd{\mathbf{48.00}}{0.46} \\

        RelCon
        & \meanstd{40.52}{0.98}
        & \meanstd{26.38}{1.33}
        & \meanstd{27.25}{0.92}
        & \meanstd{41.95}{0.41}
        & \meanstd{40.30}{0.88} \\
        \textbf{RelCon + MorphCL}
        & \meanstd{\mathbf{49.93}}{0.30}
        & \meanstd{\mathbf{50.44}}{0.75}
        & \meanstd{\mathbf{50.33}}{0.45}
        & \meanstd{\mathbf{50.30}}{0.65}
        & \meanstd{\mathbf{50.33}}{0.45} \\
        \bottomrule
    \end{tabular}
\end{table}

\subsection{MorphCL Sampling Method}

We also investigate the effects of the balancing and oversampling of MorphCL batches. We remove the balancing sampler components from MorphCL training on WISDM: (i) undersampling instead of oversampling for balance, and (ii) no balancing at all (uniform batches). Numbers are averages over the five downstream tasks, averaged across three seeds. Results pretraining on the WISDM dataset (see Table~\ref{tab:sampling_strategies_wisdm}) suggest stability of our method even when not using oversampling and balanced batches. Same as on WISDM, results on the CAPTURE-24 dataset show robustness of our method to different sampling techniques (see Table~\ref{tab:sampling_strategies_cap24}). One can see that SimCLR slightly benefits from unbalanced MorphCL batches, which we hypothesize is due to a lower likelihood that MorphCL loss gradients collide with negative-pair gradients in the SimCLR loss.

\begin{table}[!ht]
    \centering
    \tiny
    \setlength{\tabcolsep}{3pt}
    \renewcommand{\arraystretch}{1.2}
    \caption{Downstream F1 scores for different balancing strategies for the MorphCL loss pretraining on the WISDM dataset.}
    \label{tab:sampling_strategies_wisdm}
    \begin{tabular}{@{}lccc@{}}
        \toprule
        Method & Balanced + Oversampling & Balanced + Undersampling & Unbalanced \\
        \midrule
        SimCLR + MorphCL & \meanstd{50.82}{0.50} & \meanstd{50.67}{0.28} & \meanstd{50.57}{0.60} \\
        MTL + MorphCL    & \meanstd{50.53}{0.45} & \meanstd{50.42}{0.51} & \meanstd{50.28}{0.48} \\
        \bottomrule
    \end{tabular}
\end{table}

\begin{table}[!ht]
    \centering
    \tiny
    \setlength{\tabcolsep}{3pt}
    \renewcommand{\arraystretch}{1.2}
    \caption{Downstream F1 scores for different balancing strategies for the MorphCL loss pretraining on the CAPTURE-24 dataset.}
    \label{tab:sampling_strategies_cap24}
    \begin{tabular}{@{}lccc@{}}
        \toprule
        Method & Balanced + Oversampling & Balanced + Undersampling & Unbalanced \\
        \midrule
        SimCLR + MorphCL & \meanstd{50.97}{0.23} & \meanstd{51.59}{0.65} & \meanstd{52.08}{0.57} \\
        MTL + MorphCL    & \meanstd{52.12}{0.33} & \meanstd{52.31}{0.57} & \meanstd{52.24}{0.60} \\
        \bottomrule
    \end{tabular}
\end{table}

\subsection{Loss Term Weighting}

To investigate the robustness of our results presented in the main paper to different scaled MorphCL loss terms, we report sample linear probing results using three scaling factors ($\lambda \in [5, 10, 20]$) using a ViT-S backbone in combination with MTL (see Table~\ref{tab:loss_scaling}). In general, observing tracked gradient magnitudes of each individual loss, the MorphCL loss is on average around 5 to 10 order of magnitude smaller than the SSL objectives. Results in Table~\ref{tab:loss_scaling} show that unweighted summation ($\lambda=1$) of MorphCL with MTL shows no significant difference compared to weighted sums ($\lambda>1$), while inclusion of the loss term ($\lambda > 0$) shows a clear improvement. This shows the robustness of our results reported in the main paper to the scaling factor of our proposed MorphCL loss.

\begin{table}[!ht]
    \centering
    \tiny
    \setlength{\tabcolsep}{3pt}
    \renewcommand{\arraystretch}{1.2}
    \caption{Ablation on different loss scaling factors for MorphCL loss term on WISDM dataset.}
    \label{tab:loss_scaling}
    \begin{tabular}{@{}lccccc@{}}
        \toprule
         & $\lambda=0$ & $\lambda=1$ & $\lambda=5$ & $\lambda=10$ & $\lambda=20$ \\
        \midrule
                 & \meanstd{47.41}{0.22} & \meanstd{50.53}{0.45} & \meanstd{49.96}{0.99} & \meanstd{49.33}{0.66} & \meanstd{48.98}{0.84} \\
        \bottomrule
    \end{tabular}
\end{table}

\subsection{Extended Qualitative Analysis}
\label{sec:extendedqualatative}

\begin{figure}[!ht]
    \centering
    \includegraphics[width=0.9\linewidth]{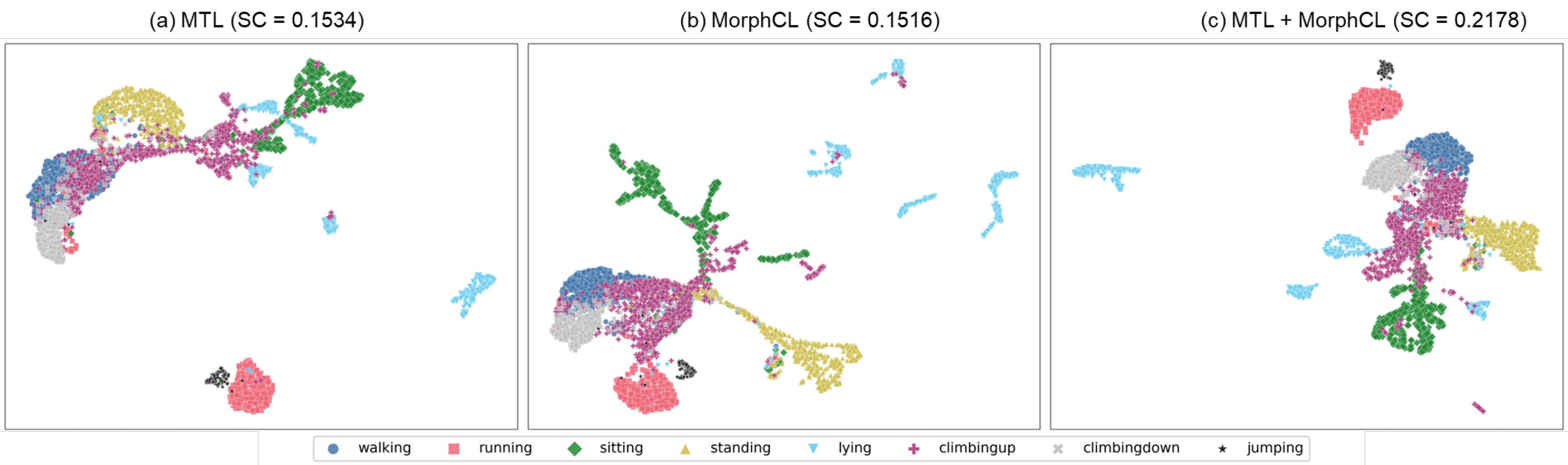}
    \caption{UMAP features and Silhouette coefficients (SC) extracted from a ViT-S pretrained on CAPTURE-24 applied to the RWHAR dataset using (i) MTL, (ii) MorphCL and (iii) MTL + MorphCL as pretext task. One can see a clear improvement when combining MorphCL with MTL with even kinematically close classes such as walking stairs ("climbingup"/"climbingdown") and walking are better separated.}
    \label{fig:umaprwhar}
\end{figure}

\begin{figure}[!ht]
    \centering
    \includegraphics[width=0.9\linewidth]{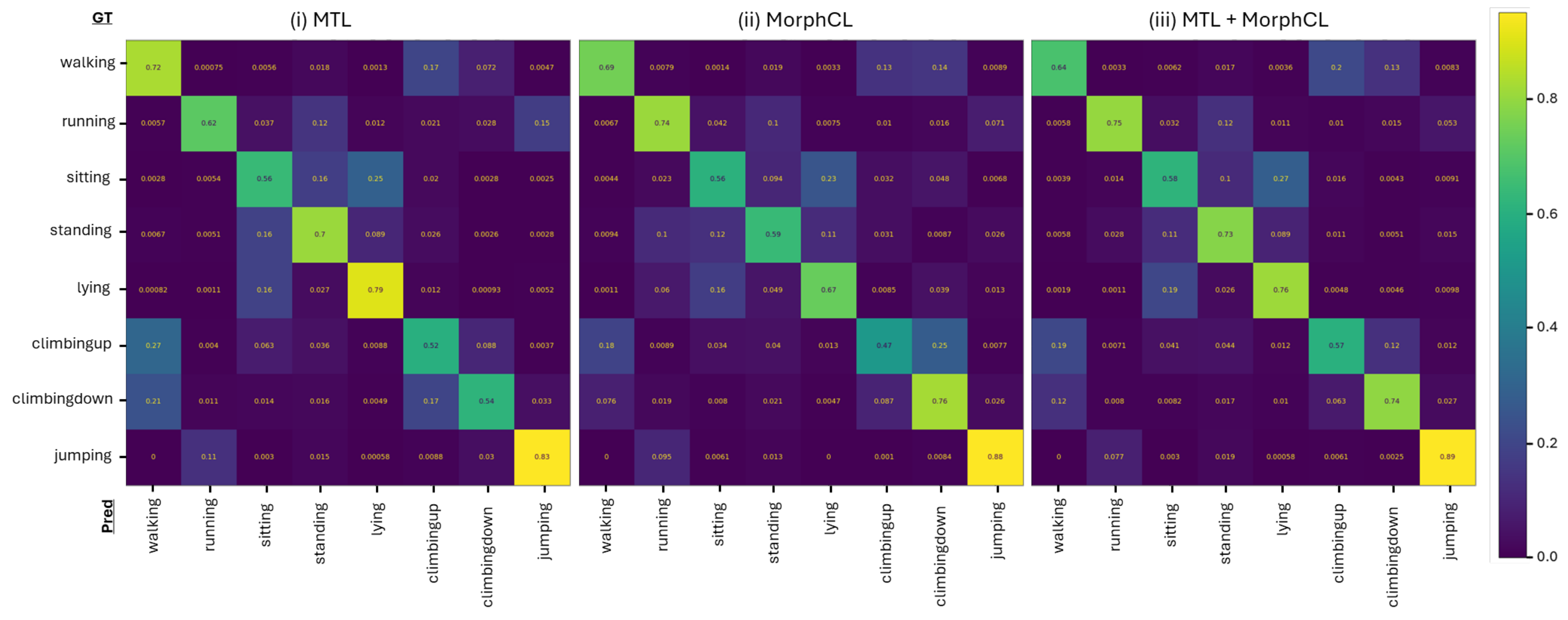}
    \caption{Linear probing results of a ViT-S pretrained on CAPTURE-24 applied to the RWHAR dataset using (i) MTL, (ii) MorphCL and (iii) MTL + MorphCL as pretext task. When combining MorphCL with MTL overall confusion across both kinematically close (e.g., "jumping" and "running") is decreased.}
    \label{fig:confrwhar}
\end{figure}

\begin{figure}[!ht]
    \centering
    \includegraphics[width=0.9\linewidth]{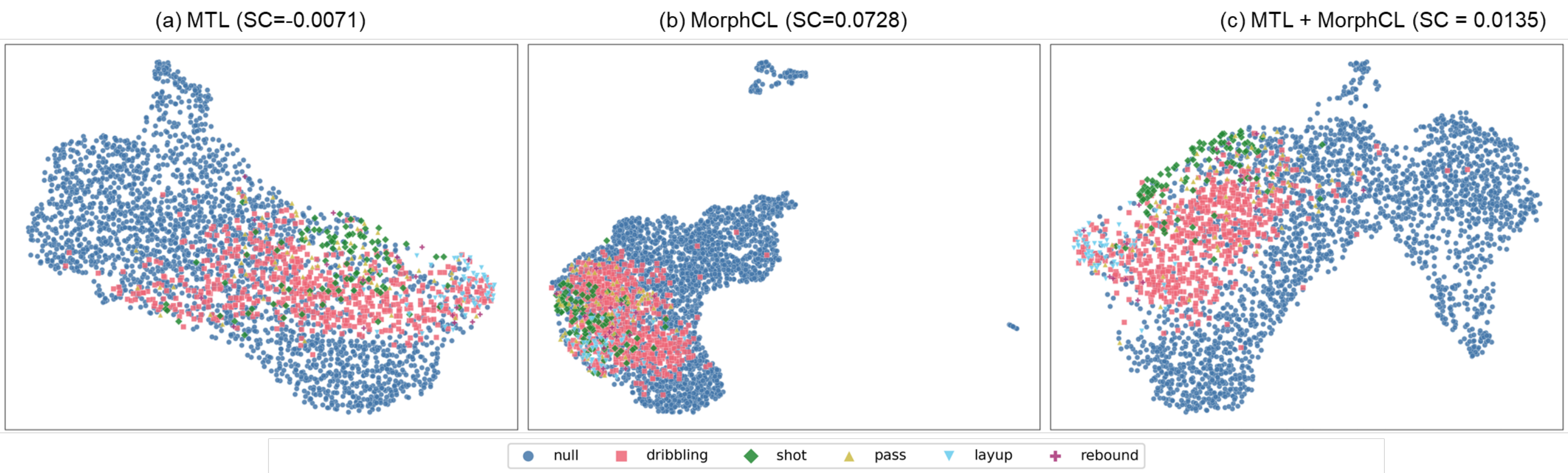}
    \caption{UMAP features and Silhouette coefficients (SC) extracted from a ViT-S pretrained on CAPTURE-24 applied to the Hang-Time dataset using (i) MTL, (ii) MorphCL and (iii) MTL + MorphCL as pretext task. Even with the presence of a dominant background class ("null"), combining MorphCL with MTL still lead to overall better separation of all classes.}
    \label{fig:umaphangtime}
\end{figure}

To underline qualitative results we provide additional plots extracted from linear probing experiments using a ViT-S pretrained on CAPTURE-24 using either the (i) MTL, (ii) MorphCL or (iii) MTL + MorphCL as loss function. We provide extracted embeddings UMAP features from sample participants of the Hang-Time (see Figure~\ref{fig:umaphangtime}) and RWHAR dataset (see Figure~\ref{fig:umaprwhar}) along with confusion matrices (see Figure~\ref{fig:confrwhar}) that underline the improved recognition performances and decreased confusion among even difficult classes. Table~\ref{tab:silhouette} further provides all computed silhouette coefficients (SC) computed on the extracted embeddings across all participants of a dataset per pretrained model. 

\begin{table}[ht]
    \centering
    \tiny
    \setlength{\tabcolsep}{3pt}
    \renewcommand{\arraystretch}{1.2}
    \caption{Silhouette coefficients (SC $\in$ [-1, 1], higher is better)  based on activity classes computed on embeddings extracted from pretrained models using each tested SSL framwork (with and without MorphCL). We report scores based on all participants in each downstream dataset using cosine similarity and excluding background classes if present. Adding MorphCL increases the mean silhouette coefficient across datasets for each evaluated configuration.}
    \label{tab:silhouette}
    \begin{tabular}{lccccccc}
        \toprule
        \textbf{Framework} & WEAR & Wetlab & PAMAP2 & Hangtime & RWHAR & WISDM & \textit{Avg} \\
        \midrule
        \multicolumn{8}{l}{\textbf{Encoder:} DeepConvLSTM} \\ 
        \midrule
        \multicolumn{8}{l}{\textbf{Pretraining:} WISDM} \\
        MTL & -0.1776 & -0.2189 & 0.0533 & -0.0471 & -0.0639 & -0.2035 & -0.1096 \\
        MTL + MorphCL & -0.2019 & -0.2025 & 0.1088 & -0.0493 & -0.0872 & -0.1682 & -0.1001 \\
        SimCLR & -0.1990 & -0.2002 & 0.0449 & -0.0742 & -0.0928 & -0.1814 & -0.1171 \\
        SimCLR + MorphCL  & -0.1906 & -0.1964 & 0.0741 & -0.0780 & -0.1096 & -0.1961 & -0.1161 \\
        REBAR & -0.2003 & -0.1809 & 0.0713 & -0.0204 & -0.1492 & -0.1862 & -0.1109 \\
        REBAR + MorphCL  & -0.1681 & -0.1947 & 0.0698 & -0.0810 & -0.1039 & -0.1723 & -0.1084 \\
        RelCon  & -0.1803 & -0.2131 & 0.0383 & -0.0598 & -0.1068 & -0.1926 & -0.1190 \\
        RelCon + MorphCL & -0.2039 & -0.2199 & 0.0648 & -0.0561 & -0.0983 & -0.1697 & -0.1139 \\
        \multicolumn{8}{l}{\textbf{Encoder:} ViT-S} \\ 
        \midrule
        \multicolumn{8}{l}{\textbf{Pretraining:} WISDM} \\ 
        MTL  & -0.1826 & -0.1274 & 0.1650 & -0.1269 & -0.0887 & -0.1309 & -0.0819 \\
        MTL + MorphCL & -0.1155 & -0.0845 & 0.2326 & -0.0444 & -0.0163 & -0.0912 & -0.0199 \\
        SimCLR & -0.0270 & -0.0906 & 0.1833 & -0.0534 & -0.0468 & -0.1614 & -0.0326 \\
        SimCLR + MorphCL & -0.0135 & -0.0966 & 0.2321 & 0.0110 & 0.0337 & -0.0983 & 0.0114 \\
        DenoisingAE  & -0.1140 & -0.1854 & 0.0271 & 0.0766 & -0.0658 & -0.1648 & -0.0710 \\
        DenoisingAE + MorphCL & -0.0022 & -0.1440 & 0.0991 & 0.0437 & -0.0098 & -0.1360 & -0.0249 \\
        MaskedAE & -0.1728 & -0.2079 & 0.0627 & 0.0208 & -0.1098 & -0.1949 & -0.1003 \\
        MaskedAE + MorphCL & -0.0236 & -0.0661 & 0.2075 & -0.0071 & 0.0444 & -0.0535 & 0.0169 \\
        REBAR & -0.1541 & -0.1383 & 0.0460 & -0.2109 & -0.0296 & -0.1546 & -0.1069 \\
        REBAR + MorphCL & -0.0429 & -0.0846 & 0.2100 & -0.0525 & 0.0208 & -0.0682 & -0.0029 \\
        RelCon & -0.2373 & -0.2158 & 0.0838 & -0.0550 & -0.0971 & -0.1789 & -0.1167 \\
        RelCon + MorphCL & -0.0739 & -0.1378 & 0.1177 & 0.0323 & -0.0612 & -0.1009 & -0.0373 \\
        \midrule
        \multicolumn{8}{l}{\textbf{Pretraining:} CAPTURE-24} \\
        MTL & -0.0484 & -0.2233 & 0.1206 & 0.0053 & -0.0972 & -0.1813 & -0.0707 \\
        MTL + MorphCL  & -0.0195 & -0.0564 & 0.3024 & -0.0436 & 0.0160 & -0.0213 & 0.0296 \\
        SimCLR & -0.0265 & -0.0611 & 0.3186 & -0.1853 & -0.0170 & -0.0252 & 0.0006 \\
        SimCLR + MorphCL & -0.0171 & -0.0498 & 0.2893 & -0.1576 & -0.0102 & -0.0140 & 0.0068 \\
        DenoisingAE & -0.0856 & -0.1759 & 0.1162 & -0.0177 & -0.1154 & -0.1202 & -0.0664 \\
        DenoisingAE + MorphCL & -0.0170 & -0.0684 & 0.3211 & -0.0333 & 0.0484 & -0.0368 & 0.0357 \\
        MaskedAE & -0.1598 & -0.2133 & 0.0468 & 0.0638 & -0.1263 & -0.2060 & -0.0991 \\
        MaskedAE + MorphCL & 0.0027 & -0.0582 & 0.2968 & -0.1234 & 0.0317 & -0.0399 & 0.0183 \\
        REBAR & -0.3077 & -0.2113 & 0.0675 & -0.2416 & -0.1708 & -0.3194 & -0.1972 \\
        REBAR + MorphCL & -0.1188 & -0.0972 & 0.2125 & -0.1885 & -0.0864 & -0.1385 & -0.0695 \\
        RelCon & -0.2234 & -0.1757 & 0.0012 & 0.0457 & -0.0166 & -0.2586 & -0.1045 \\
        RelCon + MorphCL & -0.0199 & -0.1248 & 0.2279 & -0.1514 & -0.0528 & -0.0831 & -0.0340 \\
        \bottomrule
    \end{tabular}
\end{table}

\begin{figure}[!ht]
    \centering
    \includegraphics[width=0.9\linewidth]{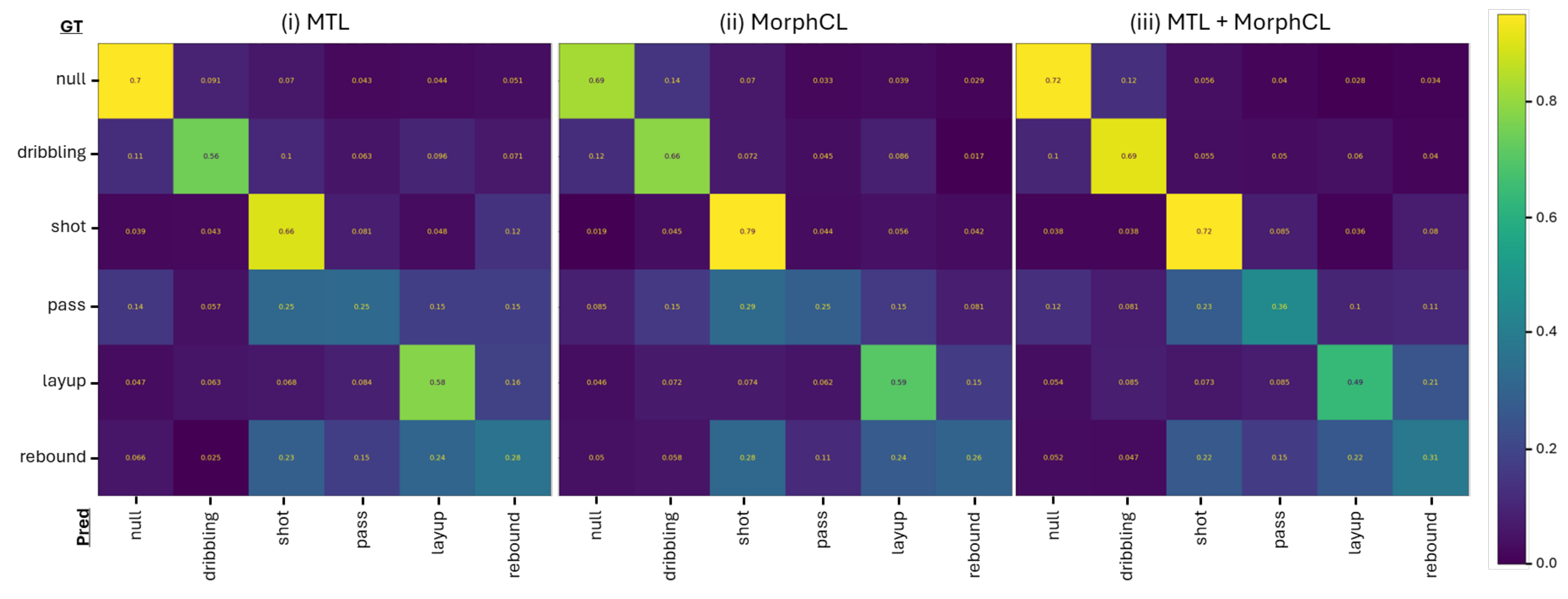}
    \caption{Linear probing results of a ViT-S pretrained on CAPTURE-24 applied to the Hang-Time dataset using (i) MTL, (ii) MorphCL and (iii) MTL + MorphCL as pretext task. One can see a clear improvement when combining MorphCL with MTL with even sporadic classes like passes being recognizable.}
    \label{fig:confhangtime}
\end{figure}

\subsection{SupCon vs. Group-based averaging}

As discussed in the introduction and related work of this paper, both~\citet{khoslaSupervisedContrastiveLearning2020} and~\citet{pillaiPaPaGeiOpenFoundation2025} rely on computing losses based on an inherent grouping among the samples during their SSL approaches. With~\citet{khoslaSupervisedContrastiveLearning2020} relying on actual class labels to define the grouping,~\citet{pillaiPaPaGeiOpenFoundation2025} use a pre-computed measure (sVRI) and group together records based on a pre-defined binning of said measure. With their methods inherently differing from our MorphCL framework they further apply different loss calculations based on the grouping. \citet{khoslaSupervisedContrastiveLearning2020} uses an instance-based formulation in which the loss is computed independently for each positive pair. For a given anchor, the denominator aggregates similarity scores over all other samples in the batch regardless of label, while the numerator isolates the similarity to a single positive instance. The resulting per-pair log-probabilities are then averaged first over all positives of each anchor, and subsequently over all valid anchors in the batch. \citet{pillaiPaPaGeiOpenFoundation2025} adopt a structurally similar per-pair formulation, computing the loss as the difference between the raw similarity of each positive pair and the log-sum-exp of similarities to all non-self samples, averaged uniformly across all positive pairs in the batch rather than first averaging within each anchor. In both cases the loss treats each positive pair independently, normalizing against the full batch. 
Table~\ref{tab:supconloss} compares results of switching our proposed loss computation with the losses as used by~\citet{khoslaSupervisedContrastiveLearning2020} and~\citet{pillaiPaPaGeiOpenFoundation2025} within our framework. One can see that our group-based loss formulation (with few exceptions) results in the on average highest F1-scores across the different downstream experiments. Note that results were computed same as the main results of this paper and are also seed-averaged across three runs.

\begin{table*}[!ht]
\centering
\tiny
\setlength{\tabcolsep}{3pt}
\renewcommand{\arraystretch}{1.2}
\caption{Ablation study on using losses as reported by~\citet{khoslaSupervisedContrastiveLearning2020} and~\citet{pillaiPaPaGeiOpenFoundation2025}. Results are averaged across the MorphCL framework itself as well as its combination with the different SSL frameworks reported in the main paper. Best results are in \textbf{bold}.}
\label{tab:supconloss}
\begin{tabular}{lccccc}
\toprule
\multirow{1}{*}{}
& \multicolumn{1}{c}{WEAR}
& \multicolumn{1}{c}{Wetlab}
& \multicolumn{1}{c}{PAMAP2}
& \multicolumn{1}{c}{Hangtime}
& \multicolumn{1}{c}{RWHAR}\\
\midrule
\multicolumn{6}{l}{\textbf{Encoder:} DeepConvLSTM; \textbf{Pretraining:} WISDM} \\
\citet{khoslaSupervisedContrastiveLearning2020}   & \meanstd{44.98}{0.34} & \meanstd{23.89}{0.27} & \meanstd{63.40}{0.43} & \meanstd{34.80}{0.42} & \meanstd{64.68}{0.71} \\
\citet{pillaiPaPaGeiOpenFoundation2025}           & \meanstd{44.94}{0.30} & \meanstd{23.96}{0.23} & \meanstd{63.34}{0.44} & \meanstd{34.59}{0.41} & \meanstd{64.25}{0.48} \\
Ours                                              & \meanstd{46.50}{0.30} & \meanstd{23.94}{0.24} & \meanstd{64.39}{0.50} & \meanstd{35.06}{0.40} & \meanstd{64.29}{0.55} \\ \midrule
\multicolumn{6}{l}{\textbf{Encoder:} ViT-S; \textbf{Pretraining:} WISDM} \\
\citet{khoslaSupervisedContrastiveLearning2020}   & \meanstd{53.20}{0.52} & \meanstd{25.75}{0.53} & \meanstd{66.77}{0.89} & \meanstd{35.71}{0.86} & \meanstd{68.18}{0.87} \\
\citet{pillaiPaPaGeiOpenFoundation2025}           & \meanstd{53.05}{0.39} & \meanstd{26.06}{0.62} & \meanstd{66.68}{1.09} & \meanstd{35.52}{0.75} & \meanstd{67.64}{1.13} \\
Ours                                              & \meanstd{53.28}{0.30} & \meanstd{26.13}{0.62} & \meanstd{65.89}{0.65} & \meanstd{36.85}{0.64} & \meanstd{68.25}{1.47} \\ \midrule
\multicolumn{6}{l}{\textbf{Encoder:} ViT-S; \textbf{Pretraining:} CAPTURE-24} \\
\citet{khoslaSupervisedContrastiveLearning2020}   & \meanstd{53.46}{0.42} & \meanstd{25.75}{0.63} & \meanstd{66.88}{0.54} & \meanstd{37.24}{0.81} & \meanstd{66.58}{0.58} \\
\citet{pillaiPaPaGeiOpenFoundation2025}           & \meanstd{53.44}{0.40} & \meanstd{25.46}{0.77} & \meanstd{66.56}{0.86} & \meanstd{37.70}{1.05} & \meanstd{66.90}{0.90} \\
Ours                                              & \meanstd{54.36}{0.34} & \meanstd{26.45}{0.78} & \meanstd{67.07}{0.46} & \meanstd{37.99}{1.11} & \meanstd{65.10}{0.97} \\
\bottomrule
\end{tabular}
\end{table*}

\end{document}

%% file: math_commands.tex
\usepackage{amsmath,amsfonts,bm}

\def\eqref#1{equation~\ref{#1}}

\def\1{\bm{1}}

\DeclareMathAlphabet{\mathsfit}{\encodingdefault}{\sfdefault}{m}{sl}
\SetMathAlphabet{\mathsfit}{bold}{\encodingdefault}{\sfdefault}{bx}{n}

